\documentclass[10pt,twocolumn,letterpaper]{article}

\usepackage{wacv}      

\definecolor{wacvblue}{rgb}{0.21,0.49,0.74}
\usepackage[pagebackref,breaklinks,colorlinks,allcolors=wacvblue]{hyperref}
\usepackage[utf8]{inputenc} 
\usepackage[T1]{fontenc}    
\usepackage{hyperref}       
\usepackage{url}            
\usepackage{booktabs}       
\usepackage{amsfonts}       
\usepackage{nicefrac}       
\usepackage{microtype}      
\usepackage{xcolor}         
\usepackage{adjustbox}
\usepackage{pifont}
\usepackage{enumitem}
\usepackage{multirow}
\usepackage{bm}
\usepackage{subcaption}
\usepackage{amsmath}
\newcommand{\cmark}{\textcolor{green!60!black}{\ding{51}}}
\newcommand{\xmark}{\textcolor{red!70!black}{\ding{55}}}  

\def\wacvPaperID{1510} 
\def\confName{WACV}
\def\confYear{2027}

\title{CoMPAS3D: A Dataset and Benchmark for Interactive Dance}

\author{Bermet Burkanova, Yasaman Etesam, Payam Jome Yazdian, Trinity Evans, \\Chuxuan Zhang, Zoe Stanley, Paige Tuttösí and Angelica Lim\thanks{Corresponding author: {\tt\small angelica@sfu.ca}} \\
Simon Fraser University, Burnaby, BC, Canada\\
{}
}

\begin{document}

\maketitle

\begin{figure}[ht]
      \centering
      \includegraphics[width=\columnwidth]{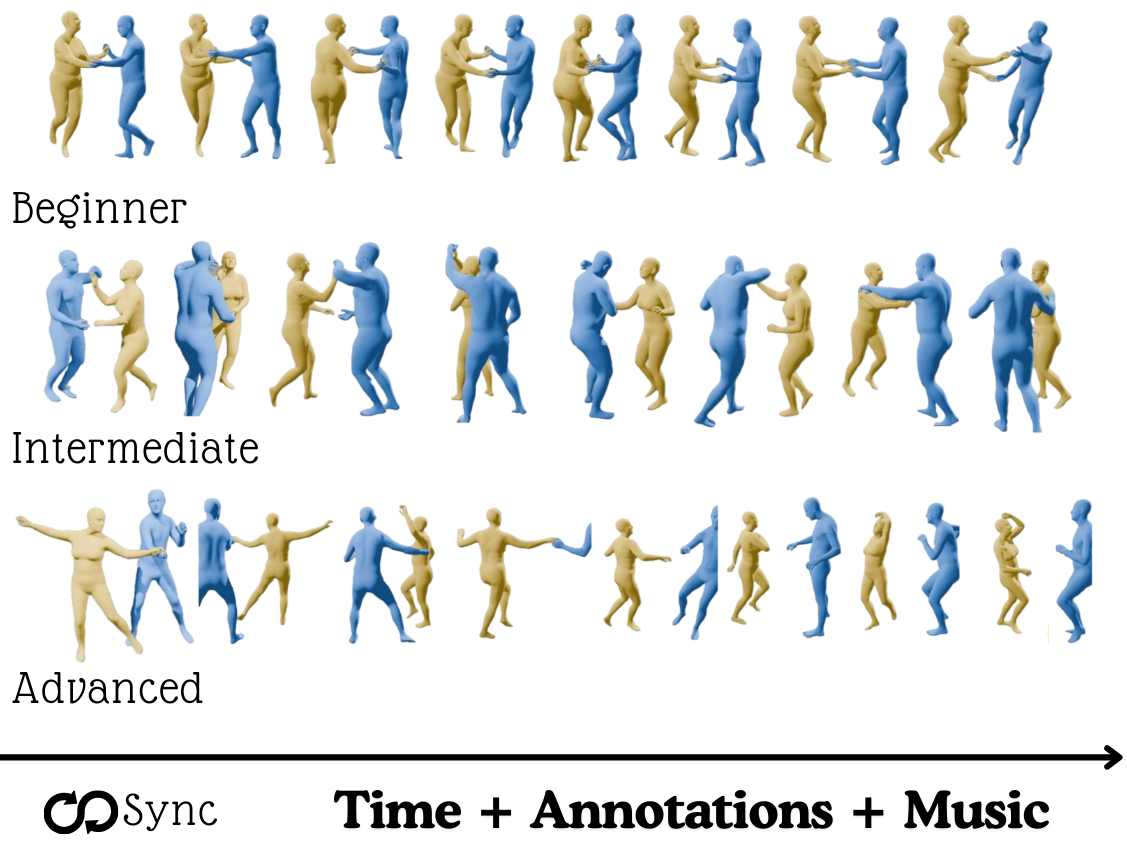}
      \caption{CoMPAS3D comprises 3 hours of improvised salsa dance with beginner (top), intermediate (middle) and professional (bottom) pairs with synchronised music and fine-grained annotations.}
      \label{Fig:teaser}
   \end{figure}

\begin{abstract}
Socially interactive humanoids aim to interact with humans through their bodies, adapting in real time to a partner's movement, intent, and abilities. This requires models that understand how bodies move as well as what their movement means. Yet, evaluation frameworks for interactive motion generation do not measure whether a generated reactive motion is valid or appropriate within a shared movement vocabulary. This gap persists for two reasons: existing evaluation frameworks rely on kinematic metrics such as FID and beat alignment that cannot measure this semantic property, and existing improvised interaction datasets lack the fine-grained move annotations to do so. Inspired by computational evaluations on interactive games such as Go or Chess, we propose Salsa, one of the world's most popular social dances, as an embodied interaction evaluation domain: it is interactive, dyadic, and governed by a move vocabulary and community-validated judging criteria covering timing, musicality, technique, difficulty, partnering, and originality. We present CoMPAS3D, a motion capture dataset of improvised partner salsa dance paired with an evaluation framework covering kinematic quality, two objective metrics (move legibility and proficiency appropriateness), and six real-world competition-based subjective dimensions. The dataset comprises 3 hours of leader-follower improvisation by 18 dancers spanning beginner, intermediate, and professional skill levels, with over 2,800 expert-annotated segments covering move types, execution errors, and stylistic elements. We define three benchmark tasks: move classification, proficiency estimation, and follower generation. Fine-tuned vision-language models achieve good performance on the objective metrics when applied to ground-truth sequences, as well as correlation with subjective expert judgments, suggesting promise as measurement tools. Applied to Duolando and InterGen, our objective metrics suggest a large gap between groundtruth and generated dances. Expert human evaluations across six judging dimensions support this same gap between generated and groundtruth motions. CoMPAS3D, annotations, benchmark code, and baseline results are publicly available at {\small{\url{https://huggingface.co/datasets/Rosie-Lab/compas3d}}}

\end{abstract}

\section{Introduction}

The long-term vision of socially interactive humanoids requires machines that can engage with humans through their bodies, adapting in real time to a partner's movement, intent, and ability level~\cite{al2009generating, dejaegher2007sensemaking}. As in Chess or Go, this problem requires generating appropriate responses to an active participant whose behavior cannot be scripted or predicted; further, the responses may carry social meaning, appropriate to the context, the partner, and the shared interaction.

Interactive motion generation research has made remarkable progress, yet its evaluation infrastructure has not kept pace. Existing metrics for generative motion measure physical realism and distribution similarity, and a recent review noted that this low-level focus has resulted in \textit{motion realism} becoming a saturated evaluation measure \cite{nagy2026reliable}. Reliable evaluations of more specialised aspects are needed, e.g., semantic alignment or emotional expression~\cite{nagy2026reliable}. 
For instance, current evaluation methods do not measure whether a generated motion is semantically valid or \textit{legible} within the shared movement vocabulary nor \textit{appropriate} to the characteristics of the interaction partner (e.g. child, adult). Interactive social motion, whether for a virtual embodied agent, robot dancing partner, or a virtual rehabilitation coach requires generated motion that is responsive to the partner's cues, legible, and adaptive to the characteristics of the human. Kinematic metrics provide limited information for this higher semantic level, just as acoustic quality metrics cannot assess whether a speaker said something sensible.

Closing the evaluation gap requires an interactive motion domain with three properties that are rarely found together: (1) improvised dyadic interaction, so the generative challenge covers the complexity of naturalistic data; (2) an externally validated evaluation scheme, so evaluation criteria are grounded in expert knowledge; and (3) multiple proficiency levels, so models can be tested for skill-appropriate generation. We identify salsa partner dance as uniquely satisfying all three. It is improvised, structured around a community-codified movement vocabulary \cite{hanna1987dance, patel2023super}, and evaluated by competition judges using established criteria covering timing, musicality, technique, difficulty, partnering/connection, and originality \cite{canadasalsacongress_rules_2024}. 


Salsa is “arguably the world’s most popular partnered social dance form”~\cite{McMains02012016}, practiced globally across a wide range of skill levels and cultural contexts. Inspired by long-standing evaluations on interactive games such as Chess \cite{ruoss2024amortized} or Go \cite{44806}, we present the embodied CoMPAS3D (Complex Multi-Level Person-Interaction Annotated Salsa Dataset), a motion capture dataset of improvised salsa duets together with a three-level evaluation framework for social motion. The dataset comprises 3~hours of leader-follower improvisation by 18~dancers across beginner, intermediate, and professional proficiency levels, with over 2,800 expert-annotated move segments covering move types, execution errors, and stylistic elements. The evaluation framework combines kinematic metrics (FID, diversity, beat alignment~\cite{T2MGPT, siyao2022bailando}), two objective metrics measuring move \textit{legibility} and proficiency \textit{appropriateness} via fine-tuned vision-language models trained on our expert annotations, and subjective metrics via six real-world competition-based subjective dimensions rated by expert human evaluators~\cite{canadasalsacongress_rules_2024}. The objective metrics operate in two stages: we first validate that fine-tuned vision language models (VLMs) accurately classify moves and proficiency on ground-truth sequences (Table~\ref{tab:classification}), then apply them to generated sequences to assess whether follower motion is legible and proficiency-appropriate.

We apply the framework to two state-of-the-art reaction generation methods: Duolando, a dance-specific follower generation model~\cite{siyao2024duolando}, and InterGen~\cite{liang2024intergen}, a general human-human interaction method that generates both agents simultaneously. Our contributions are:


\begin{itemize}
    \item CoMPAS3D, to the best of our knowledge, the first \textbf{openly available} (1) improvised partner dance dataset with (2) expert-annotated dance move transcriptions (3) spanning three proficiency levels
 \item Benchmarks of three pretrained and fine-tuned vision-language models (VLMs) on two novel tasks enabled by our dataset: dance move classification and proficiency level estimation

  \item Objective evaluation metrics for reactive dance generation  \cite{siyao2024duolando, liang2024intergen} using the move classifier and proficiency estimators, showing that generative methods still fail to produce legible or proficiency-appropriate motions

 \item Subjective evaluation metrics for reactive dance generation based on community-based dance competition metrics, with 3 new metrics compared to prior work in dance generation
 
\end{itemize}

\section{Related Work}


\begin{table*}[t]
\centering
\renewcommand{\arraystretch}{1.3}
\setlength{\tabcolsep}{8pt}
\caption[Comparison of publicly available interactive partner dance datasets.]{Comparison of publicly available dance datasets capturing human-human interaction (HHI). $\bar{T}$/s represents the average duration per sequence in seconds. \textbf{Pairs/Genre} highlights the depth of coverage within a single movement vocabulary: DD100 captures 0.5 pairs per genre across 10 ballroom styles, and InterDance's pairs per genre is unknown across 15 genres, whereas CoMPAS3D dedicates all 9 pairs to a single genre, similar to a richly annotated dataset in English rather than shallow coverage across multiple languages.}
\resizebox{\textwidth}{!}{
\begin{tabular}{lccccccccccccc}
\toprule
\textbf{Dataset} & \textbf{\# Participants} & \textbf{\# Genres} & \textbf{Pairs/Genre} & \textbf{Audio} & \textbf{Experience} & \textbf{$\bar{T}$/s} & \textbf{$T$} & \textbf{\# Frames} & \textbf{Improvised} & \textbf{Mocap} & \textbf{Annotated} & \textbf{Representation} \\ \midrule
ReMoCap & 9 & n/a & n/a & \xmark & Pro & Varies & 2.04h & 275.7K & \xmark & \xmark & \xmark & 3D Skeletons \\ 
ExPI & 4 & 1 & 2 & \xmark & Pro & 10.4 & 0.33h & 30K & \xmark & \cmark & \cmark & 3D Meshes \\ 
InterHuman & 30 & n/a & n/a & \xmark & Pro & 10 & 6.56h & 1.7M & \xmark & \cmark & \cmark & SMPL \\
InterDance & Unknown & 15 & Unknown & \cmark & Pro & 142.7 & 3.93h & 1.7M & \cmark & \cmark & \xmark & SMPL-X \\ 
DD100 & 10 & 10 & 0.5 & \cmark & Pro & 70.2 & 1.95h & 210K & \cmark & \cmark & \xmark & SMPL-X \\ 
CoMPAS3D & 18 & 1 & 9 & \cmark & Beg, Int., Pro & 150 & 3.0h & 1.3M & \cmark & \cmark & \cmark & SMPL-X \\ \bottomrule
\end{tabular}
}
\label{tab:interaction_datasets}
\end{table*}

\textbf{Partner Dance and Social Motion Datasets.} Partner dance datasets offer a promising source of long-term physical interaction with structured movement vocabularies. ExPI \cite{guo2022multi} captures Lindy Hop dancing with 3D body poses and shapes, DD100 \cite{siyao2024duolando} provides 117 minutes of music-synchronized SMPL-X data from professional dance pairs, and InterDance \cite{li2024interdance} offers 3.93 hours of optical motion capture across 15 genres. Synergy and Synchrony \cite{maluleke2024synergy} presents an in-the-wild video dataset of Swing dancing focused on future motion prediction, but relies on estimated 3D poses from video rather than motion capture and provides no move annotations. The salsa dataset of Senecal et al. \cite{senecal2018motion, senecal2019classification} include skill-level variation, and the dance dataset of Gupta et al. \cite{gupta2025mdd} include annotations, but are not publicly available for machine learning research at the time of writing. We summarize these datasets in Table \ref{tab:interaction_datasets}. While these datasets offer valuable resources, they differ from real-world embodied communication in three key ways: they often rely on choreographed rather than spontaneous performances, capture only professional dancers rather than a diversity of skill levels, and lack fine-grained annotations of moves, errors, or styling. While many general human-human interaction datasets exist \cite{xu2024interx, van2016spatio, yin2023hi4d, liu2019ntu, fieraru2023reconstructing, mehta2018single, peng2023mi, ghosh2024remos, xu2024regennet}, this paper focuses specifically on partner dance, given the general ability to provide scoring and judging criteria. 

\textbf{Vision-Language Models for Motion Understanding.} Recent vision-language models such as Qwen2-VL~\cite{wang2024qwen2vl} demonstrate strong video understanding capabilities, enabling fine-tuning for domain-specific classification tasks. However, general VLMs lack the domain-specific vocabulary needed to assess move legibility or proficiency appropriateness in social dance. We investigate whether fine-tuning on our expert move annotations enables evaluation of whether motion is legible within the salsa movement vocabulary and appropriate to the target proficiency level. This parallels the use of a trained move-legality classifier to evaluate gameplay in Chess or Go, which relies on the availability of move-level annotations. While action recognition models have explored motion classification in single-person settings~\cite{liu2019ntu}, they do not address the dyadic, context-dependent nature of social motion evaluation.

\textbf{Summary.} Existing open datasets for human-human interaction largely focus on short-term, scripted performances by professional participants, providing no move-level annotations or proficiency variation. Existing evaluation frameworks rely on kinematic metrics that measure physical realism but cannot assess whether a follower's motion is legible within the shared movement vocabulary or appropriate to the skill level of the interaction. As shown in Table~\ref{tab:interaction_datasets}, to the best of our knowledge, CoMPAS3D provides the first publicly available improvised partner dance dataset with expert move annotations spanning three proficiency levels, enabling two new objective evaluation dimensions: move legibility and proficiency appropriateness. The focus on salsa also grounds human studies in real-world competition criteria. Like AI chess research, which favors depth in a single well-defined ruleset over breadth across many boardgames, we concentrate on one activity and ruleset, allowing us to verify move validity and apply specific judging criteria.

\section{The CoMPAS3D Dataset}

\begin{figure*}[ht]
      \centering
      \includegraphics[width=\textwidth]{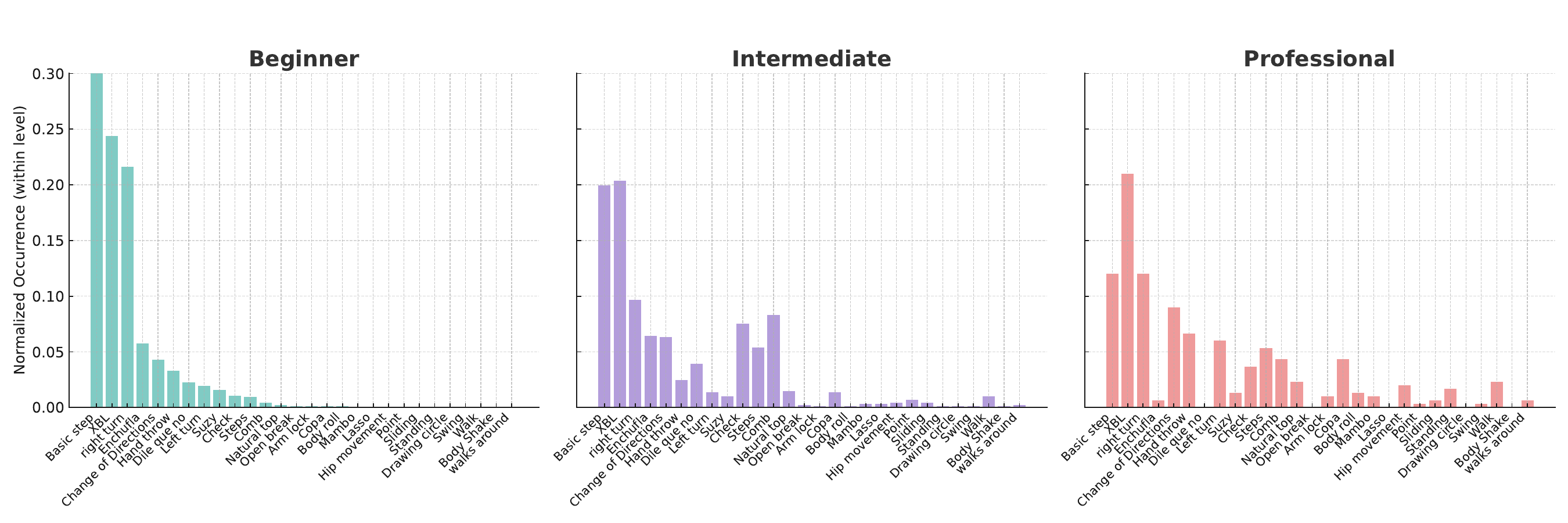}
      \caption{Distribution over the 30 move classes (sorted by beginner move frequency) in CoMPAS3D for beginner, intermediate and pro pairs. Beginners tend to primarily use ``basic step'', which professionals use less. Pros use a wider variety of moves, e.g. left turns and copa.}
       \label{Fig:move_stats}
\end{figure*}

To support the study of improvised, naturalistic nonverbal communication in physical interactions, we introduce \textbf{CoMPAS3D} (Complex Multi-Level Person-Interaction Annotated Salsa Dataset), a large-scale motion capture dataset of salsa duet dances. CoMPAS3D,  \textit{compas} meaning rhythm in Spanish, consists of over 3.0 hours of improvised leader-follower interactions performed by 18 participants spanning beginner, intermediate, and professional skill levels. Each recording captures long-duration sequences of continuous social improvisation, annotated at the frame level for move types, stylistic variations, and execution errors. The dataset includes synchronized audio recordings, high-fidelity 3D motion data and SMPL-X parametric body model fits~\cite{SMPL-X:2019}, enabling detailed analysis and modeling of embodied conversational dynamics across skill levels. The focus on a single genre (salsa) is a deliberate design choice motivated by the need for depth within a shared movement vocabulary and ruleset. Following this depth-over-breadth principle, CoMPAS3D dedicates all pairs to a single genre rather than spreading participants thinly across styles.

\textbf{Participants}. CoMPAS3D includes 18 participants, forming 9 dancing pairs. Participants were recruited from a university salsa club, community dance groups, and professional dance schools. To capture variation in fluency and style, we recruited 3 pairs each for the following salsa experience levels: beginner (3–12 months), intermediate (1–3 years), and professional (>3 years). This diversity enables the study of movement improvisation and fluency across a wide proficiency spectrum. This study was approved by the university ethics board. Each participant was compensated \$100 for 1 hour of study participation time and provided informed consent for their anonymized motion capture data release prior to data collection.


\textbf{Collection Setup}. Recordings were conducted in a controlled studio environment using a Vicon motion capture system equipped with 20 cameras operating at 120 frames per second. Each dancer wore 53 markers following the Vicon ``FrontWaist'' marker set. Improvisation sessions used four salsa music tracks (90–105 beats per minute) chosen to vary in mood and tempo. Each pair performed two improvised takes per song, each lasting approximately 2.5 minutes, resulting in a total of 72 sequences.

\textbf{Data Representation}. We release the dataset to facilitate a wide range of machine learning and animation applications. Each sequence includes 55-joint SMPL-X~\cite{SMPL-X:2019} human body joint trajectories and fitted parameters (.npz), as well as visualizations with synchronized music tracks (.mp4). We also provide ELAN annotation files (.txt) aligned frame-by-frame with the motion data.
   
\textbf{Annotation}. Approximately half of the recorded sequences (2803 segments) were annotated manually by an expert salsa dancer with 15 years of salsa dance experience and competition judging experience. Salsa moves are performed in 8-beat cycles, where the leader typically provides a signal in the early part of the cycle, and the follower completes the move by the end of the 8th beat. Therefore, each sequence was split into 8-beat segments and annotated. Each annotation contains a primary move category selected from among 30 move categories; these move categories are listed and explained in the Appendix (in Supplementary Materials). Annotations also include common execution errors (e.g., off-beat errors, mixed signals), and presence of styling (e.g., arm styling, hip accents, annotated as ``lady styling'' or ``man styling''). In addition to each broad move category, a detailed description of the move, including hand holds and secondary combinations, is provided for each segment. This detailed annotation effort using the ELAN software~\cite{elan2026} required over 120 hours. As a quality check, 5\% of the move annotations were randomly selected and verified with a second expert salsa dancer with 14 years of salsa dance experience, producing an agreement score of 0.752 using Cohen's Kappa (substantial agreement). Half the sequences remain unannotated, offering a clean set for future evaluation. 

\textbf{Analysis.} We analyzed the annotations to better understand the differences between the populations of dancers in our dataset. In Fig.~\ref{Fig:move_stats}, we compare the move distributions between beginner, intermediate and professional dancers. We noticed that professionals employ a wider variety of moves and use fewer ``basic steps''. An analysis of the styling annotations indicated that professionals execute 54.5 styling moves per performance, followed by intermediates with 12.9 styling moves per performance, and beginners, who incorporate 5.1 styling moves per performance. The most common error is ``off beat'' suggesting that multimodal information including music is important in detecting errors. Another error is unclear signals from the leader, sometimes resulting in a failed move.

\begin{figure*}[]
      \centering
      \includegraphics[width=\textwidth]{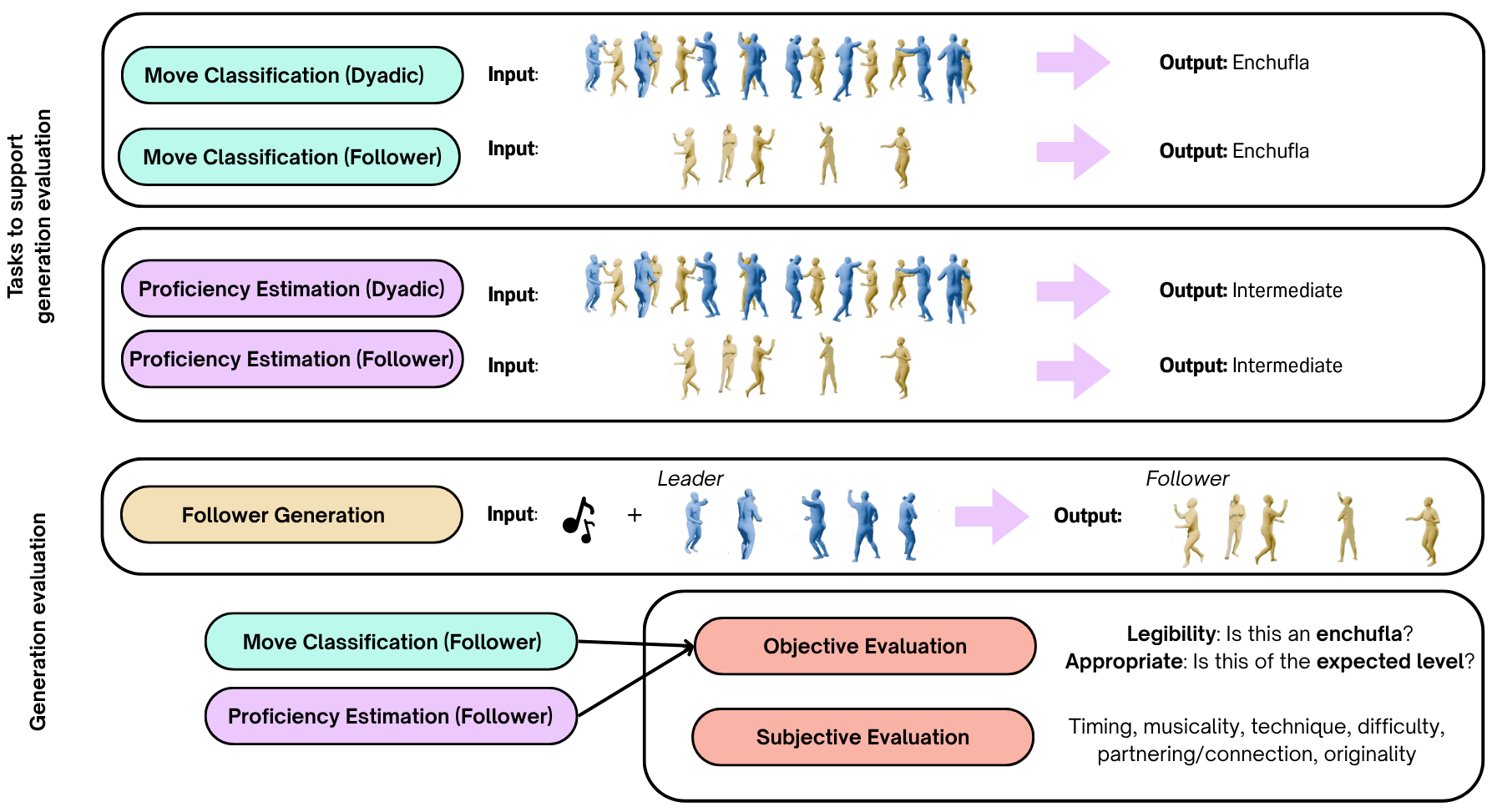}
      \caption{Proposed benchmark tasks for the CoMPAS3D dataset: (1) move classification (dyadic and on solo follower moves), (2) proficiency estimation and (3) follower generation. Objective evaluation of follower dance generation uses (1) and (2).}
      \label{Fig:benchmark_tasks}
   \end{figure*}

\section{Benchmark Tasks}
\label{sec:tasks}

We define three benchmark tasks on CoMPAS3D. The tasks progress from understanding to generation: move classification tests whether a model can identify what move is being performed; proficiency estimation tests whether it can assess the skill level of a dancer or dancing pair; and follower generation tests whether a model can produce follower motion that is both legible and proficiency-appropriate in response to a leader's cues. Fig.~\ref{Fig:benchmark_tasks} summarizes the inputs and outputs for each task.

\label{sec:task_move}
\paragraph{Move Classification.} Move classification identifies the salsa move being performed from a motion sequence. Just as automatic speech recognition requires a corpus of transcribed spontaneous speech, this task is made possible by the expert move annotations in CoMPAS3D. Each annotated 8-beat segment is labeled with one of 30 move categories (see Appendix); the task takes a motion sequence as input and predicts the move label. We note that this task differs from end-to-end transcription, as segmentation is treated as given. Automatic segmentation of continuous dance into move boundaries is an open problem and a direction for future work, analogous to early keyword spotting approaches in speech recognition prior to full end-to-end automatic speech recognition (ASR).

Beyond serving as a standalone benchmark, move classification is also used to assess move legibility in generated follower sequences, as described in Section~\ref{sec:eval_metrics}. For this task, we used the 11 moves in the training set which had at least 20 instances: \textit{Basic Step}, \textit{Change of Direction}, \textit{Check}, \textit{Comb}, \textit{Copa}, \textit{Dile que no}, \textit{Hand Throw}, \textit{Right Turn}, \textit{Enchufla}, \textit{Left Turn}, and \textit{XBL}. An \textit{Other} class was also added.

\label{sec:task_proficiency}

\paragraph{Proficiency Estimation.} Similar to fluency assessment in second language acquisition, proficiency estimation identifies the skill level of a dancing pair from their motion. The task takes a motion sequence as input and predicts one of three proficiency levels: beginner, intermediate, or professional. As shown in Fig.~\ref{Fig:move_stats}, proficiency is reflected in a combination of move vocabulary, styling density, and timing accuracy.
\begin{table*}[]
  \centering
  \caption{Classification results on CoMPAS3D. Move classification (left) reports accuracy (Acc.) and macro-averaged weighted F1 for single-person (follower only) and dyadic (leader + follower) settings. Proficiency estimation (right) identifies skill level from motion. Underline indicates the model and metric used for follower generation evaluation.}
  \label{tab:classification}
  \vspace{4pt}
  \small
  \setlength{\tabcolsep}{3pt}
  \hspace{-70pt}
  \begin{minipage}[t]{0.55\linewidth}
    \centering
    \textit{Move Classification}\\[6pt]
    \begin{tabular}{l|l cc cc}
      \toprule
      & & \multicolumn{2}{c}{Single-Person} & \multicolumn{2}{c}{Dyadic} \\
      \cmidrule(r){3-4} \cmidrule(l){5-6}
      &Method & Acc. & F1 & Acc. & F1 \\
      \midrule
      \multirow{3}{*}{\rotatebox[origin=c]{90}{FT}} 
      &Qwen2.5-VL & {61.85} & {44.36} & {69.87} & {49.30} \\
      &LLaVA-NeXT-Video & {60.24} & {44.16} & {73.36} & {\textbf{51.24}} \\
      &InternVL3 & {\textbf{69.88}} & {\textbf{\underline{53.55}}} & {\textbf{74.24}} & {51.09} \\
      \midrule
      \multirow{3}{*}{\rotatebox[origin=c]{90}{0-shot}}
      &Qwen2.5-VL & {23.69} & {5.76} & {3.93} & {1.71} \\
      &LLaVA-NeXT-Video & {21.29} & {2.92} & {23.14} & {3.13} \\
      &InternVL3 & {\textbf{29.32}} & {\textbf{6.31}} & {\textbf{27.95}} & {\textbf{7.74}} \\
      \bottomrule
    \end{tabular}
  \end{minipage}
  \hspace{-5pt}
  \small
    \setlength{\tabcolsep}{3pt}
  \begin{minipage}[t]{0.32\linewidth}
    \centering
    \textit{Proficiency Estimation}\\[6pt]
    \begin{tabular}{l|l cc cc}
      \toprule
      & & \multicolumn{2}{c}{Single-Person} & \multicolumn{2}{c}{Dyadic} \\
      \cmidrule(r){3-4} \cmidrule(l){5-6}
      &Method & Acc. & F1 & Acc. & F1 \\
      \midrule
      \multirow{3}{*}{\rotatebox[origin=c]{90}{FT}} 
      &Qwen2.5-VL & {51.51} & {\textbf{\underline{51.64}}} & {\textbf{84.63}} & {\textbf{84.53}} \\
      &LLaVA & {\textbf{52.82}} & {43.74} & {83.14} & {82.64} \\
      &InternVL3 & {42.20} & {42.62} & {75.91} & {75.20} \\
      \midrule
      \multirow{3}{*}{\rotatebox[origin=c]{90}{0-shot}}
      &Qwen2.5-VL & {\textbf{39.72}} & {\textbf{29.54}} & {\textbf{48.72}} & {\textbf{38.87}} \\
      &LLaVA & {34.42} & {21.70} & {34.14} & {16.97} \\
      &InternVL3 & {32.19} & {16.24} & {39.41} & {29.10} \\
      \bottomrule
    \end{tabular}
  \end{minipage}
\end{table*}
\label{sec:task_generation}
\paragraph{Follower Generation.} Analogous to generating a coordinated next move in response to a partner's move, follower generation predicts the follower's motion given the leader's motion and the shared music. This is the primary generative task on CoMPAS3D. Unlike solo dance generation, which conditions only on music and proficiency, follower generation requires the model to interpret the leader's cues and produce a response that is legible within the move vocabulary and appropriate to the target proficiency level. This task is evaluated at three levels: kinematic metrics assess physical realism; the move classifier assesses move legibility; and the proficiency estimator assesses proficiency appropriateness.

\section{Evaluation Metrics}
\label{sec:eval_metrics}

For the follower generation task (Section~\ref{sec:task_generation}), we evaluate 
using kinematic metrics from prior work, along with proposed objective and subjective metrics.

\paragraph{Kinematic metrics.} For individual follower quality, we report standard metrics from previous work: 
Fréchet Inception Distance in kinematic and graphical feature spaces 
($\text{FID}_k$, $\text{FID}_g$) and diversity ($\text{Div}_k$, 
$\text{Div}_g$)~\cite{siyao2024duolando}.
For 
interaction quality we use cross-distance FID and diversity 
($\text{FID}_{cd}$, $\text{Div}_{cd}$). For rhythmic consistency we 
report Beat Echo Degree (BED) for leader--follower synchrony and Beat 
Alignment Score (BAS) for motion--music alignment~\cite{siyao2022bailando, 
siyao2024duolando}. We note that salsa relies on syncopated (off-beat) rhythms~\cite{10.3389/fnhum.2016.00064}, and therefore BAS closest to the groundtruth BAS is preferable.

\paragraph{Objective metrics.} We propose measuring generated follower motion  \textit{legibility} through \textit{move classification F1-score} and \textit{appropriateness} to partner characteristics through \textit{proficiency estimation F1-score}. We compute these on generated follower sequences using the best performing fine-tuned VLMs from Table~\ref{tab:classification}, underlined. These metrics assess whether generated motion is legible within the salsa movement vocabulary and appropriate to the target proficiency.

\paragraph{Subjective metrics.} We conducted a human evaluation study with three expert salsa dancers with a mean of 9.0 years of experience (range: 5–12 years). They rated short video clips of groundtruth dance (GT), InterGen, and Duolando on a 5-point Likert scale across the six competition dimensions~\cite{canadasalsacongress_rules_2024}: 
timing, musicality, technique, difficulty, partner coordination, and originality. Details on their definitions are provided in the Appendix. Prior duet human studies \cite{liang2024intergen} only asked participants to rate on three axes: motion quality, music-motion alignment, and partner coordination. We additionally ran the identical protocol with a larger pool of 31 non-expert dancers recruited through Prolific. These participants had general dance experience (median 3 years) but mostly outside salsa; only 4 of 31 reported any salsa experience. All studies were approved by the university research ethics board.

\section{Benchmark Experiments}
\label{sec:experiments}

We present results for the three benchmark tasks defined in Section~\ref{sec:tasks}, evaluated with standard kinematic metrics and newly proposed metrics described in Sec.~\ref{sec:eval_metrics}. For move classification and proficiency estimation, we report results from fine-tuned vision-language models in single-person and dyadic settings. For follower generation, we evaluate Duolando and InterGen (which have publicly available code) through both objective and subjective measures, with the move classifier from Sec.~\ref{sec:task_move} applied to generated sequences.

\subsection{Move Classification}
\label{sec:exp_move}

We fine-tuned Qwen2.5-VL~\cite{wang2024qwen2vl}, 
LLaVA-NeXT-Video~\cite{li2024llavanext}, and InternVL3~\cite{zhu2025internvl3} on 
the CoMPAS3D expert annotations to classify move types from rendered motion sequence 
videos, in both single-person (follower only) and dyadic (leader + follower) settings. As shown in Table~\ref{tab:classification}, 
dyadic classification consistently outperformed single-person follower classification across fine-tuned models, suggesting that leader motion carries information about the move being performed. This also suggests that follower generation can be scored on single-person motion only, to avoid the influence of groundtruth leader input on the metric. Finally, note that given the substantial class imbalance in the dataset, the F1-score provides a more reliable evaluation metric than accuracy. This is evidenced by the zero-shot InternVL3 move classifier on Duolando, which predicts the ``Basic Step'' label more than 70\% of the time yet achieves a relatively high accuracy  (Table~\ref{tab:contextual}); see the Appendix for confusion matrices.


\subsection{Proficiency Estimation}
\label{sec:proficiency_estimation}

Proficiency estimation identifies the skill level of a single follower dancer or dancing pair from their motion sequence, using the same VLM fine-tuning setup as Sec.~\ref{sec:exp_move}. 
As shown in Table~\ref{tab:classification}, fine-tuned models achieved good dyadic 
proficiency estimation accuracy (84.63\% for Qwen2.5-VL, 83.14\% for LLaVA-NeXT-Video), 
with a large gap over 0-shot performance. Single-person proficiency estimation is harder, suggesting that the interaction dynamics between leader and follower 
carry important proficiency cues beyond what is visible in the follower alone. The 
proficiency estimator trained in this section is subsequently applied to generated 
follower sequences in Sec.~\ref{sec:exp_generation} to assess whether generative 
methods produce motion at the correct proficiency level.

\subsection{Follower Generation}

\newcommand{\et}[2]{$#1_{{\scriptstyle\pm#2}}$}
\begin{table*}[t]
  \centering
  \caption{Quantitative comparison on CoMPAS3D. We include ground truth as reference and two generative baselines. We present solo, interactive, and motion--music alignment metrics. $\pm$ denotes a 95\% bootstrap confidence interval over 100 runs. Arrows indicate higher ($\uparrow$), lower ($\downarrow$) or ($\rightarrow$) closer to groundtruth is better. Among \emph{generative} methods (excluding ground truth), the best value in each column is shown in \textbf{bold}.}
  \label{tab:main_quant}
  \resizebox{\textwidth}{!}{%
  \setlength{\tabcolsep}{1.4mm}%
  \scriptsize
  \begin{tabular}{l cccc ccc c}
    \toprule
     & \multicolumn{4}{c}{Solo Metrics} & \multicolumn{3}{c}{Interactive Metrics} & \multicolumn{1}{c}{Rhythmic} \\
    \cmidrule(r){2-5} \cmidrule(lr){6-8} \cmidrule(l){9-9}
    Method
      & FID$_k$($\downarrow$) & FID$_g$($\downarrow$) & Div$_k$($\uparrow$) & Div$_g$($\uparrow$)
      & FID$_{cd}$($\downarrow$) & Div$_{cd}$($\uparrow$) & BED($\uparrow$)
      & BAS($\rightarrow$) \\
    \midrule
    Groundtruth
      & \et{0.00}{.00} & \et{0.00}{.00} & $8.74$ & $0.81$
      & \et{0.00}{.00} & $2.01$ & $0.42$
      & $0.16$ \\
    \midrule
    InterGen
      & \et{24.58}{.75} & \bm{\et{13.57}{.45}} & \bm{$8.59$} & $0.72$
      & \bm{\et{18.62}{.28}} & \bm{$2.07$} & \bm{$0.32$}
      & \bm{$0.20$} \\
    Duolando
      & \bm{\et{12.46}{.45}} & \et{15.54}{.25} & $7.96$ & \bm{$0.78$}
      & \et{20.14}{.23} & $1.93$ & $0.30$
      & $0.22$ \\
    \bottomrule
  \end{tabular}%
  }
\end{table*}

\label{sec:exp_generation}

\paragraph{Kinematic evaluation.}
We evaluated two state-of-the-art follower generation methods fine-tuned on CoMPAS3D: Duolando~\cite{siyao2024duolando}, pre-trained on latin dance data, and InterGen~\cite{liang2024intergen}, a general human-human interaction model. The task involves predicting a follower's motion sequence given the groundtruth leader motion, and music. Kinematic results are shown in Table~\ref{tab:main_quant}.

\begin{table}[t]
  \centering
  \caption{Objective legibility (using InternVL3) and proficiency appropriateness (using Qwen2.5-VL) evaluations (in \textbf{bold}) on generated follower motions on CoMPAS3D. Ground truth from Table \ref{tab:classification} is included as reference.}
  \label{tab:contextual}
  \setlength{\tabcolsep}{1.4mm}
  \small
  \begin{tabular}{l|l cccc}
    \toprule
    & & \multicolumn{2}{c}{Legibility $\uparrow$} & \multicolumn{2}{c}{Appropriateness $\uparrow$} \\
      \cmidrule(r){3-4} \cmidrule(l){5-6}
      &Method & Move Acc. & Move F1 & Profic. Acc. & Profic. F1 \\
    \midrule
    & Groundtruth & 69.88 & 53.55 & 51.51 & 51.64 \\
    \midrule
    \multirow{2}{*}{\rotatebox[origin=c]{90}{FT}} 

   &  InterGen    & 18.86 & \textbf{7.69} & 49.77 & \textbf{46.55} \\
    & Duolando    & 14.47 & 6.70 & 45.47 & 39.30 \\
    \midrule
    \multirow{2}{*}{\rotatebox[origin=c]{90}{0-shot}} 
     &  InterGen    & 14.47 & 4.09 & 32.22 & 18.00 \\
    & Duolando    & 25.00 & 4.86 & 33.57 & 19.80 \\
    \bottomrule
  \end{tabular}
\end{table}

\paragraph{Objective evaluation.} We applied the best-performing move classifier (InternVL3) and best proficiency estimator (Qwen-2.5-VL) from Sections~\ref{sec:exp_move} and~\ref{sec:proficiency_estimation} to follower sequences generated by Duolando and InterGen, using the single-person setting with ground-truth leader motion, and music as input for Duolando. For InterGen, which jointly generates both agents, we overwrote the ground-truth leader motion throughout the denoising process and retain only the generated follower sequence for evaluation. Results are shown in Table~\ref{tab:contextual}. Both InterGen and Duolando scored well below ground truth on move legibility, and similarly low on proficiency appropriateness scores. We observed that both Move F1 and Proficiency F1 reflected a slight preference for \emph{InterGen} as (a) generating more legible motions, and (b) able to generate sequences that better match the proficiency of the leader (see videos in the Supp. Mat. for visualizations). This suggests that these metrics can be complementary to the kinematic evaluation results. 

\begin{figure*}[t]
    \centering
    \begin{minipage}[b]{0.49\textwidth}
        \centering
        \includegraphics[width=\textwidth]{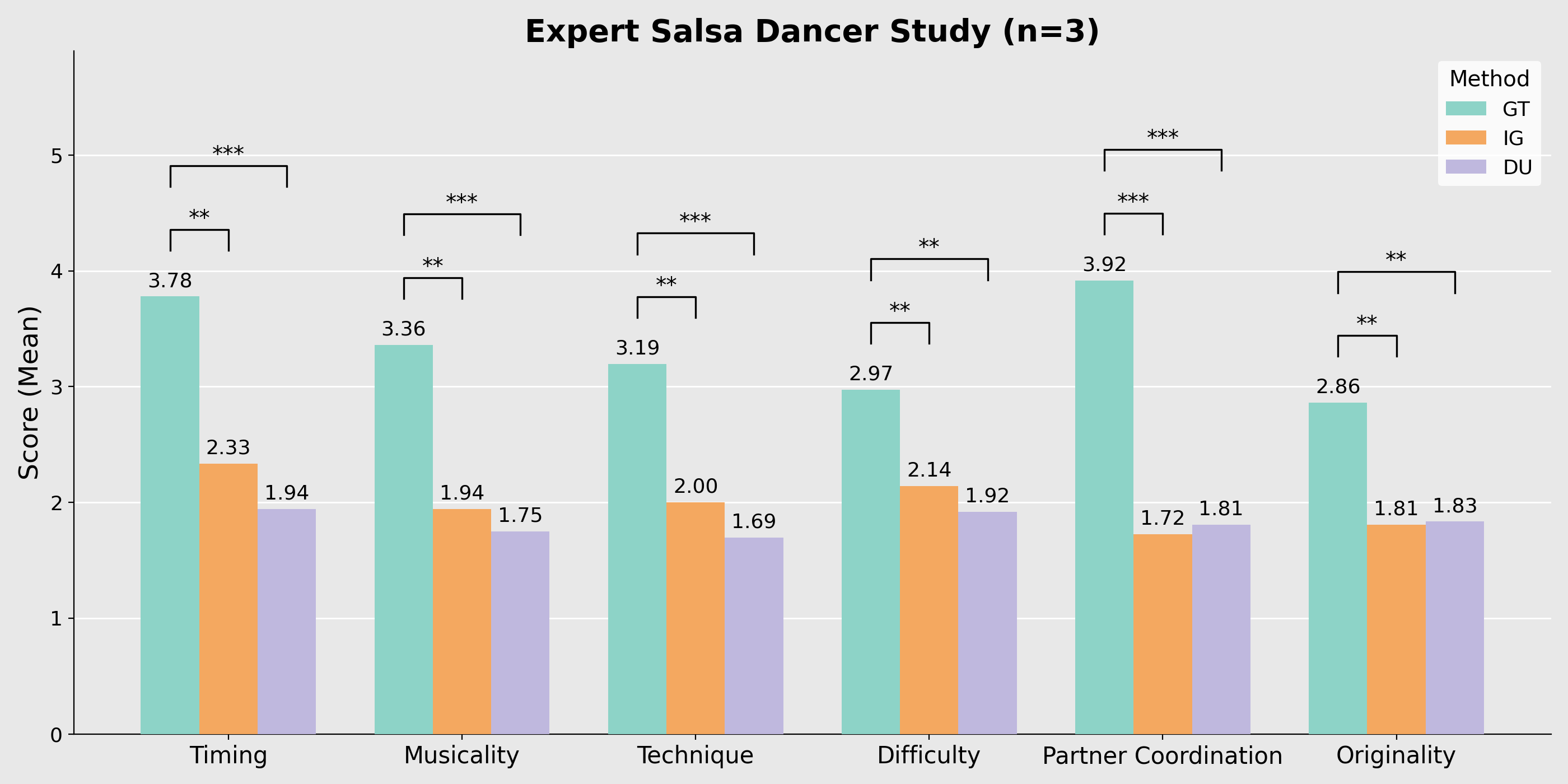}
    \end{minipage}
    \hfill
    \begin{minipage}[b]{0.49\textwidth}
        \centering
        \includegraphics[width=\textwidth]{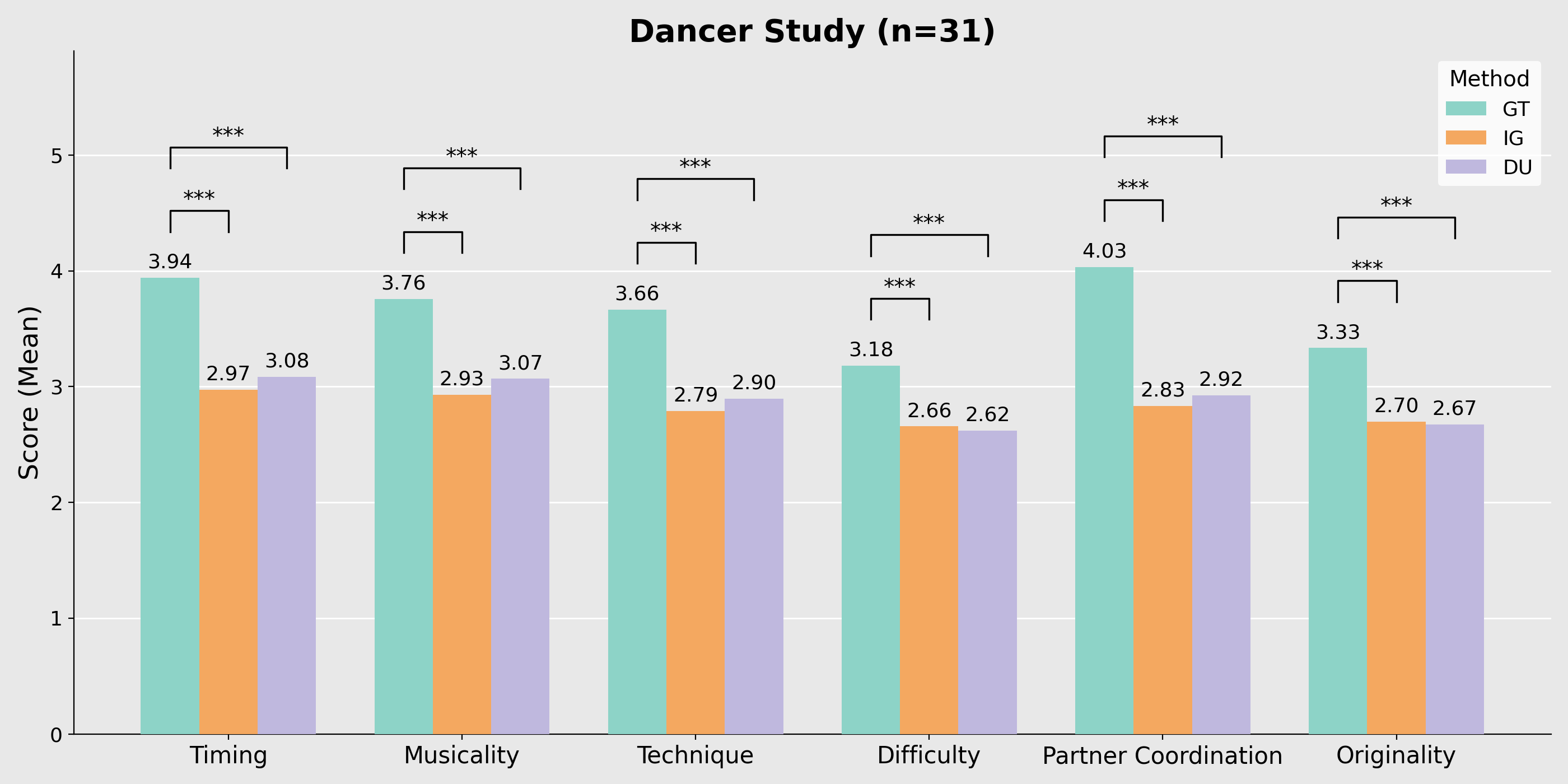}
    \end{minipage}
    \caption{Human evaluation of generated salsa dance sequences across six subjective
    dimensions, rated by (left) three expert salsa dancers (mean 9.0 years of experience,
    range 5--12) and (right) 31 non-expert participants. Bars show mean ratings (1--5 Likert
    scale) for ground truth (GT), InterGen (IG), and Duolando (DU). $^{*}p<0.05$, $^{**}p<0.01$, $^{***}p<0.001$}
    \label{fig:human_study}
\end{figure*}

\paragraph{Subjective evaluation.} We first report inter-rater agreement of our three salsa experts using a two-way mixed-effects model, computing the intra-class correlation coefficient (ICC) of expert ratings of the stimulus set. Inter-rater reliability was acceptable to good across dimensions (ICC(C,k) = 0.47–0.76, all p < 0.05), as shown in Table \ref{tab:icc_agreement}. Given this agreement, we averaged the three experts' ratings for each clip into a single consensus score per method, then compared GT to InterGen and GT to Duolando using paired two-sided Wilcoxon signed-rank tests on these per-clip consensus scores (n = 12 paired clips). For the non-expert study, we computed each participant's mean rating per method across the 12 trials, then compared GT to InterGen and GT to Duolando using paired two-sided Wilcoxon signed-rank tests on these participant-level means (n = 31). Among both experts and non-experts, we found that ground truth (GT) dances were rated significantly higher than both InterGen (IG) and Duolando (DU) across all 6 dimensions (Figure \ref{fig:human_study}).

\textbf{Do experts and non-experts evaluate similarly?} We used paired Wilcoxon tests across the 12 shared clips. Salsa experts rated both InterGen and Duolando significantly lower than non-experts on every dimension (p<0.05 to p<0.001). This suggests non-experts are more lenient toward generated salsa, likely because detecting technical flaws requires domain expertise, supporting our use of expert judgment as the primary evaluation.

\begin{table}[t]
  \centering
  \caption{Expert inter-rater agreement (ICC). Two-way mixed-effects, consistency, average measures across three expert salsa dance annotators.}
  \label{tab:icc_agreement}
  \vspace{4pt}
  \small
  \setlength{\tabcolsep}{3pt}
  \begin{tabular}{l cc}
    \toprule
    Dimension & ICC(C,k) & 95\% CI \\
    \midrule
    Timing               & 0.48$^{*}$            & [0.10, 0.72] \\
    Musicality           & 0.60$^{**}$           & [0.30, 0.78] \\
    Technique            & 0.47$^{*}$            & [0.08, 0.71] \\
    Difficulty           & 0.56$^{**}$           & [0.23, 0.76] \\
    Partner Coord.       & 0.76$^{**}$           & [0.58, 0.87] \\
    Originality          & 0.55$^{**}$           & [0.22, 0.76] \\
    Rank                 & \textbf{0.80}$^{**}$  & [0.66, 0.89] \\
    \bottomrule
    \multicolumn{3}{l}{\footnotesize $^{*}\,p < .05$\quad $^{**}\,p < .01$}
  \end{tabular}
\end{table}

\begin{figure}[t]
  \centering
  \includegraphics[width=\linewidth]{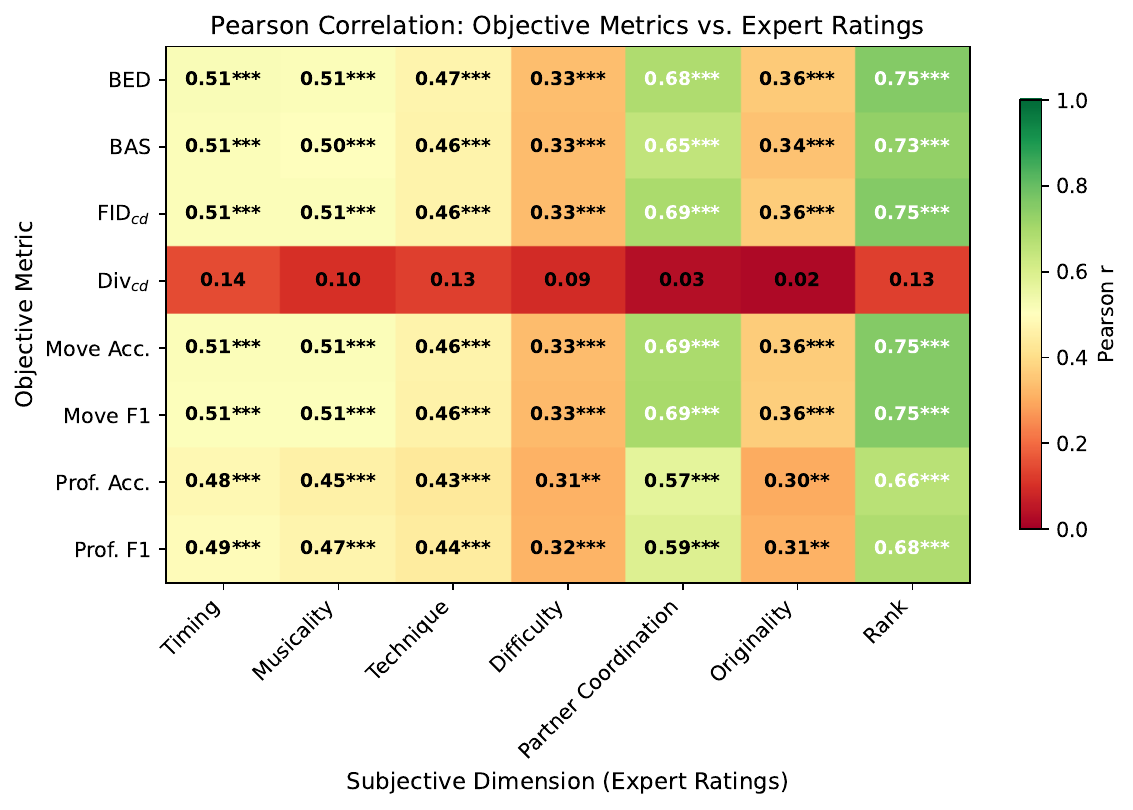}
  \caption{Correlation between objective and subjective measures.}
  \label{fig:heatmap}

\end{figure}

\textbf{How well do expert judgments correlate with our objective metrics?} We report this in Fig.~\ref{fig:heatmap}. Each heatmap cell reflects a Pearson correlation between an objective metric and expert subjective ratings, pooled across three experts, twelve trials, and three methods (GT, InterGen, Duolando) for 108 paired data points per cell. All metrics were set such that higher indicated better performance: raw values for BED, Div$_cd$, Move Accuracy, Move F1, Proficiency Accuracy, and Proficiency F1; negated values for FID$_cd$; negative distance from groundtruth's BAS value for BAS; and negated values for Rank, since rank 1 indicated the best performance.

As reported in Fig. \ref{fig:heatmap}, we found that our legibility metric (Move F1) correlated well with expert judgments of partner coordination and overall rank. We noticed that the diversity score did not correlate strongly with any subjective dimension; variety in movement may not be sufficient on its own to reflect coordinated salsa dancing. We observed that \emph{difficulty} and \emph{originality} were the weakest-correlating subjective dimensions across most metrics. We suspect this may be because difficulty is tied to the use of specific moves (Fig. 2); future work could examine how move usage relates to proficiency level, and how originality can be operationalized as a metric. We also found that proficiency accuracy correlated with expert judgments, though more weakly than the move-based metrics. This may be expected since proficiency accuracy aims to measure skill-level consistency between the two dancers, rather than perceived quality directly.

\section{Discussion and Limitations}
CoMPAS3D supports the development of socially interactive embodied agents capable of improvising dyadic interactions, as well as computational models of interpersonal synchrony~\cite{georgescu2020reduced}. It is currently limited to a single genre and 9 dancing pairs; future work can broaden coverage to other salsa variants, additional genres, and mixed-proficiency pairings. Related open directions include automatic move segmentation and transcription, move-sequence similarity metrics analogous to those used to compare game move sequences, contact and haptic signal annotations, and evaluation targets for \textit{originality} and \emph{difficulty}, which showed the weakest correlation with current metrics.


Two further limitations concern the methods evaluated on CoMPAS3D rather than the dataset itself. First, while the imperfect dances improvised by beginners are useful to train models for proficiency classification, the generative methods tested may produce non-professional dance due to the inclusion of beginner data; future work can include development of models that condition on proficiency, toward the ability to mimic different levels of dance. Second, the sampling rate of the tested VLMs may only capture general move validity rather than fine-grained errors, a limitation we expect to lessen as more efficient video models become available for use as dance move classifiers. While we do not claim that the metrics or dataset transfer directly to other movement tasks or partner dance forms, given differences in movement vocabulary, algorithms advanced through this dataset may still inform work on other judged reactive dances or on broader human-human interaction tasks lacking such criteria.

\section{Conclusion}
We introduced CoMPAS3D, a richly annotated dataset of improvised salsa duets across 
three proficiency levels. We provide objective and subjective metrics that move past kinematic scoring: legibility and proficiency appropriateness, along with subjective community-validated judging criteria, suggest that Duolando and InterGen both generate follower motion that scores well below ground truth, providing an open area of research for interactive dance generation.



\section{Acknowledgments}
We would like to thank Giorgio Becherini and Dr. Michael Black for their assistance in MOSH conversion to SMPL-X format. We also thank Ahmet Tasel and Jim Su for their help in learning the motion capture process and initial discussions. This work would also not be possible without support from the Rajan Family.

    \bibliographystyle{plain}
    \bibliography{references}

{
    \small
    \begin{thebibliography}{29}

    \bibitem[3]{appendixref-canadasalsacongress}
    Canada Salsa \& Bachata Congress.
    \newblock Rules, judging criteria \& definitions.
    \newblock \url{https://www.canadasalsacongress.com/rules}, 2026.

    \bibitem[5]{appendixref-dettmers2023qlora}
    Tim Dettmers, Artidoro Pagnoni, Ari Holtzman, and Luke Zettlemoyer.
    \newblock Qlora: Efficient finetuning of quantized llms.
    \newblock \emph{Advances in neural information processing systems},
      36:10088--10115, 2023.

    \bibitem[13]{appendixref-hu2022lora}
    Edward~J Hu, Yelong Shen, Phillip Wallis, Zeyuan Allen-Zhu, Yuanzhi Li,
      Shean Wang, Liang Wang, and Weizhu Chen.
    \newblock Lora: Low-rank adaptation of large language models.
    \newblock \emph{Proceedings of the 10th International Conference on
      Learning Representations (ICLR)}, 1(2):3, 2022.

    \bibitem[25]{appendixref-salsadictionary}
    Salsa is Good.
    \newblock Salsa dancing dictionary.
    \newblock \url{https://www.salsaisgood.com/dictionary/Salsa_dictionary.htm},
      n.d.
    \newblock Accessed: 2025-04-07.

    \bibitem[29]{appendixref-simpson2019theorizing}
    Rebecca Simpson-Litke and Chris Stover.
    \newblock Theorizing fundamental music/dance interactions in salsa.
    \newblock \emph{Music Theory Spectrum}, 41(1):74--103, 2019.

    \end{thebibliography}
}

\appendix

\clearpage
\setcounter{page}{1}
\section{CoMPAS3D: Additional Dataset Details}

The CoMPAS3D dataset is comprised of 72 salsa duet dances of 2.5min each. Each of the 9 pairs performed two takes for each of 4 songs, resulting in 8 takes each. The details on each pair, their public annotations, and the test set is in Table \ref{tab:pair_splits}. For proficiency estimation, Pair 6, Pair 8, and Pair 9 are held-out as the test set.

\begin{table}[h]
\centering
\begin{tabular}{|l|l|l|l|}
\hline
\textbf{Pair} & \textbf{Proficiency} & \textbf{Annotations} & \textbf{Test Set} \\
\hline
Pair 1 & Beginner     & 100\% & Song1\_Take1 \\
Pair 2 & Intermediate & 100\% & Song1\_Take2 \\
Pair 3 & Beginner     & 100\% & Song2\_Take1 \\
Pair 4 & Intermediate & 100\% & Song2\_Take2 \\
Pair 5 & Professional & 50\%  & Song3\_Take1 \\
Pair 6 & Intermediate & n/a   & Song3\_Take2 \\
Pair 7 & Professional & n/a   & Song4\_Take1 \\
Pair 8 & Beginner     & n/a   & Song4\_Take2 \\
Pair 9 & Professional & n/a   & Song1\_Take1 \\
\hline
\end{tabular}
\caption{Pair proficiency levels, public annotations and corresponding sequences held out for testing for generative models.}
\label{tab:pair_splits}
\end{table}

 \begin{figure*}[]
      \centering
      \includegraphics[width=\textwidth]{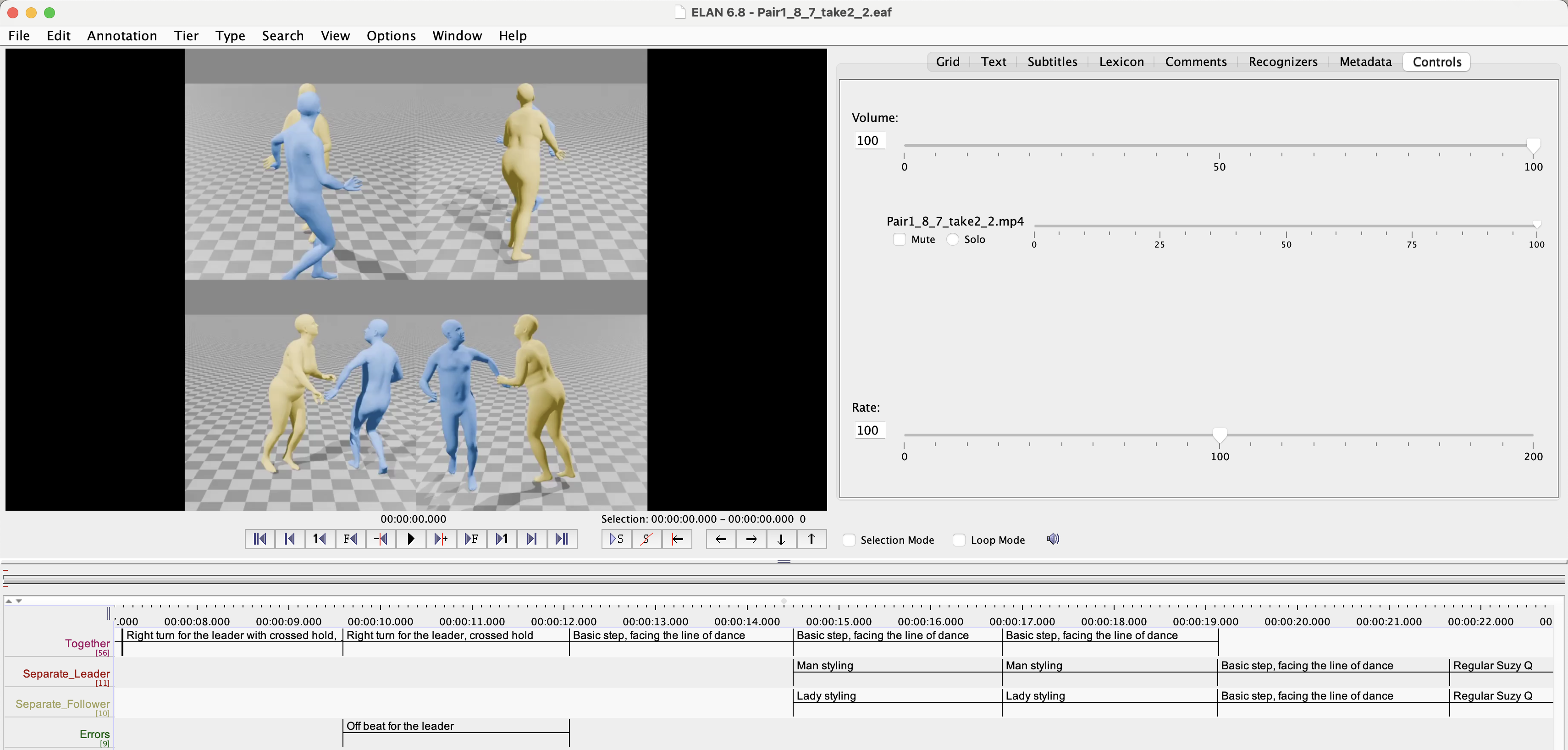}
      \caption[ELAN - annotation tool]{ELAN annotation tool used for segmenting and labeling dance moves in animated SMPL-X representation files. The annotation includes four tracks: Together – when dancers execute the move as a pair; Separate\_Leader – when the leader dances solo or adds "Man Styling" to the base move; Separate\_Follower – when the follower dances solo or incorporates "Lady Styling"; and Errors – for marking mistakes.}
      \label{Fig:elan}
   \end{figure*}

\begin{figure*}
    \centering
    \includegraphics[width=\textwidth]{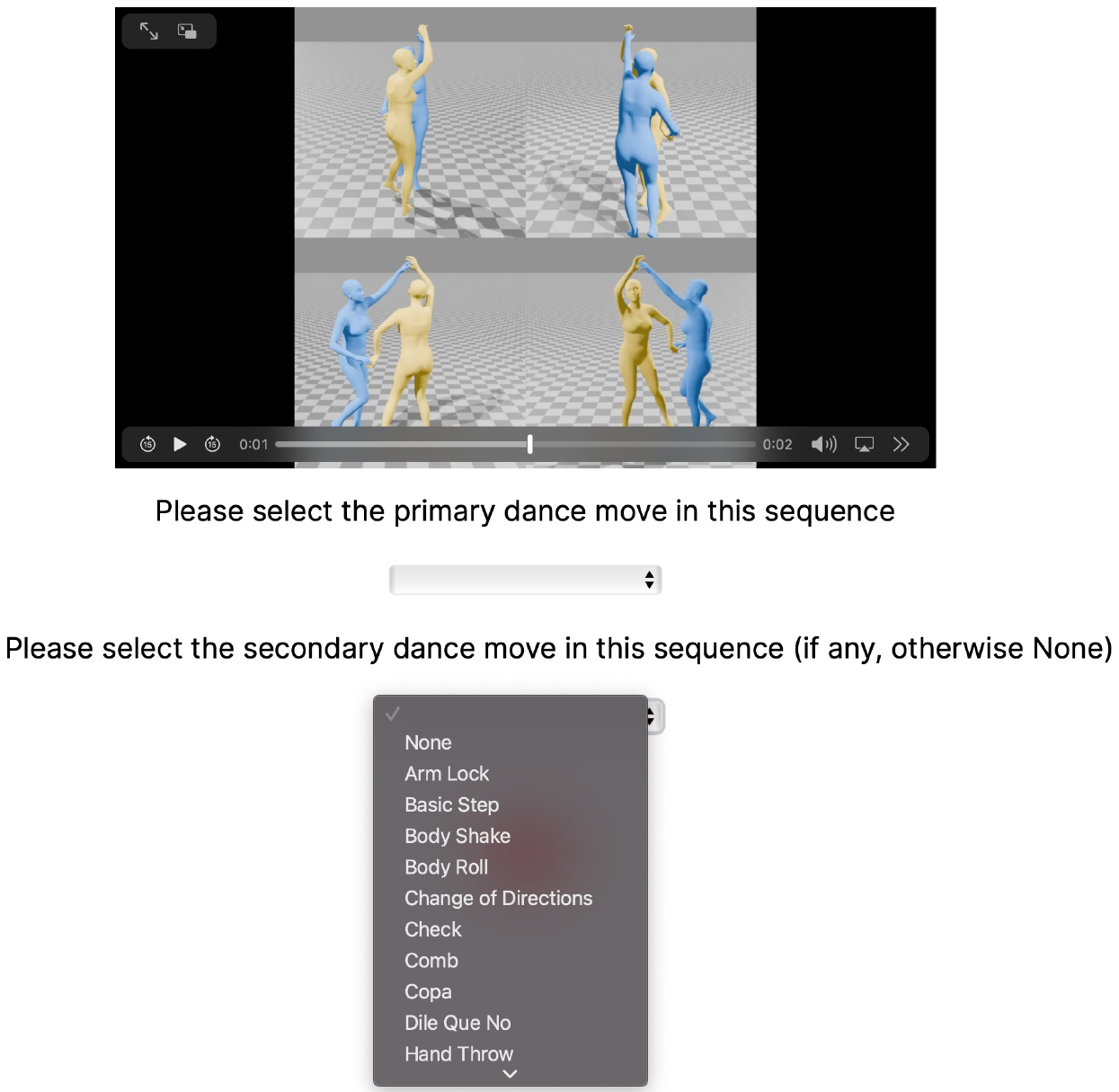}
    \caption{The annotation validation task of the salsa move identification built in Gorilla. Each trial presented a short video clip of a base move, followed by two dropdown menus prompting the second annotator to label the primary and secondary moves observed.}
    \label{fig: gorilla}
\end{figure*}

\subsection{Annotation Tool}
We utilized the ELAN annotation tool (Figure~\ref{Fig:elan}) to facilitate precise temporal and semantic labeling of the captured dance sequences. SMPL-X representations were manually synchronized with the musical tracks using the witness camera audiovisual footage, generating video files imported into ELAN. We created four annotation tracks: paired move labels, individual dancer move and styling annotations, and error classification.

\subsection{Segmentation}

Frame-accurate segmentation was achieved through rhythmic alignment based on the clave pattern, a fundamental rhythmic structure in salsa \cite{simpson2019theorizing}. The clave pattern, characterized by alternating bars of three and two beats (2-3 or 3-2), provides the dance's temporal framework. Segmentation involved marking the start and end frames of each 8-count dance sequence, typically corresponding to a complete dance move, based solely on the musical rhythm. 

\subsection{Annotation}
\begin{table*}[t]
\centering
\small
\begin{tabular}{|l|l|p{8cm}|}
\hline
\textbf{Name} & \textbf{Category} & \textbf{Description} \\ \hline
Arm lock & Move & A locking arm movement often used to create tension or highlight transitions. \\ \hline
Basic step & Move & Fundamental salsa step with variations including side, cross-back, and back basic steps. \\ \hline
Body shake & Move & A rapid shaking movement emphasizing torso dynamics. \\ \hline
Body roll & Move & A fluid, wave-like motion passing through the body. \\ \hline
Change of Directions & Move & Transition step involving directional changes, including position swaps. \\ \hline
Check & Move & A checking step used to halt or redirect movement. \\ \hline
Comb & Move & A styling-influenced move where the hand is combed over the head. \\ \hline
Copa & Move & A pivoting movement redirecting the follower after a forward step. \\ \hline
Dile que no & Move & A foundational salsa move, translating to ``tell her no.'' \\ \hline
Hand throw & Move & A dramatic throwing motion of one or both hands. \\ \hline
Right turn & Move & A clockwise rotational turn performed by the dancer. \\ \hline
Drawing circle & Move & Circular motion with hands or body to accentuate movement. \\ \hline
Enchufla & Move & Salsa turn pattern where partners switch places. \\ \hline
Walks around & Move & Continuous walking around a partner, often in a circular path. \\ \hline
Suzy Q & Move & Classic salsa footwork emphasizing rhythm and flair. \\ \hline
Hip movement & Move & Emphasized hip motion often synchronized with the rhythm. \\ \hline
Kicks & Move & Kicking action integrated within footwork patterns. \\ \hline
Lasso & Move & Overhead arm motion resembling lassoing. \\ \hline
Natural top & Move & Continuous circular motion performed with a partner. \\ \hline
Left turn & Move & A counterclockwise turn executed by the dancer. \\ \hline
Mambo & Move & Latin dance step characterized by forward and backward movements. \\ \hline
Open break & Move & A breaking step where partners create distance. \\ \hline
Point & Move & Pointing gesture typically with feet or hands. \\ \hline
Sliding & Move & Smooth gliding motion across the floor. \\ \hline
Standing & Move & Stationary stance often used for resets or transitions. \\ \hline
Steps & Move & General term for footwork elements. \\ \hline
Swing & Move & Rhythmic swinging motion involving torso or arms. \\ \hline
Walk & Move & Basic locomotion step in any direction. \\ \hline
XBL (Cross Body Lead) & Move & Core salsa move where the follower is led across the leader. \\ \hline
Indescribable & Move & Complex or ambiguous movements not fitting other categories. \\ \hline
Markers Swap issue & Move & Technical artifact caused by marker misalignment. \\ \hline
Lady styling & Styling & Feminine aesthetic enhancements involving hands, hips, and posture. \\ \hline
Man styling & Styling & Masculine aesthetic embellishments emphasizing strength and rhythm. \\ \hline
Misinterpreted signal & Error & Occurs when the follower misunderstands the leader's cue. \\ \hline
Misstep & Error & Incorrect foot placement deviating from the intended movement. \\ \hline
Mixed signals & Error & Conflicting cues from the leader resulting in follower confusion. \\ \hline
Off beat & Error & Deviation from the musical rhythm during execution. \\ \hline
\end{tabular}

\caption[Comprehensive Overview of Move, Styling, and Error Annotations in the CoMPAS3D Dataset.]{Comprehensive Overview of Move, Styling, and Error Annotations in the CoMPAS3D Dataset. This table categorizes the various elements annotated during the dataset creation process, specifying whether each element pertains to a dance move, styling, or error classification.}
\label{tab:annotation_overview}
\end{table*}
\par
\textbf{Moves.} Dance move annotations were derived from expert knowledge and standardized salsa terminology \cite{salsadictionary}. Each segmented sequence was labeled with base moves and their variations, compiled from a 20-entry dictionary (Table \ref{tab:annotation_overview}). This dictionary, based on external resources and expert additions, defined moves with base names and descriptive add-ons. For instance, a sequence could be labeled `cross body lead' followed by `follower's right turn with normal open hold', specifying the base move, follower action, and hand hold. Move complexity included simultaneous or sequential execution of multiple base moves within an 8-count cycle. To derive the primary move class from a detailed annotation, the move class, e.g. used for the classification task, was determined (Table \ref{tab:annotation_overview}) using the first four words of the detailed annotation.

\par
\textbf{Styling.} Styling annotations captured `man styling' and `lady styling', which are aesthetic embellishments of base moves through hand, foot, hip, head, shoulder, or full-body accessorization. These were classified into `no styling' (standard execution), `lady styling' (feminine embellishments), and `man styling' (masculine embellishments). These stylings, including balance, posture, locomotion, timing, body isolation, and partner connection, were annotated to analyze role-specific stylistic variations.
\par
\textbf{Errors.} Five error classes were defined: `no error', `misinterpreted signal' (leader cue misunderstanding), `misstep' (incorrect foot placement), `mixed signals' (conflicting cues), and `off beat' (deviation from musical rhythm). For example, a `Mixed signals and failed move' occurred during a `cross body lead with left (inside) crossed hold and hand change' at 00:01:56.510 - 00:01:58.860 for the second pair, second song, first take (Pair2\_8\_7\_take2\_1), where leader hesitation and an ambiguous hand movement led to follower confusion and a subsequent `copa' move. These error annotations aim to support analysis of skill levels, non-verbal communication, and identifying undesirable dance patterns.

\subsection{Music}

To capture a diverse range of couple dance dynamics, we selected 4 popular musical pieces with varying beats per minute (BPM), tempi, and musical moods (Table \ref{tab:song_info}). The music is copyrighted, with all rights remaining with the original performers. The release of the music in our dataset within .mp4 video files was reviewed by the university copyright office and deemed fair use.

\begin{table*}[]
\centering
\caption{Songs used in the CoMPAS3D dataset with artist names and tempos.}
\begin{tabular}{|c|l|l|c|}
\hline
\textbf{Song} & \textbf{Artist} & \textbf{Title} & \textbf{Tempo (BPM)} \\
\hline
Song 1 & Tito Rojas & \textit{Lo que te queda} & 90 \\
Song 2 & Louie Ramirez, Ray de La Paz & \textit{Lluvia} & 105 \\
Song 3 & Leoni Torres & \textit{Idilio} & 95 \\
Song 4 & Johnny Ventura & \textit{Dilema} & 93 \\
\hline
\end{tabular}
\label{tab:song_info}
\end{table*}

\section{Human Evaluation Study Details}
\label{sec:human_study}

\begin{figure*}[h]
    \centering
    \hspace*{-2cm}
    \includegraphics[width=1.3\textwidth]{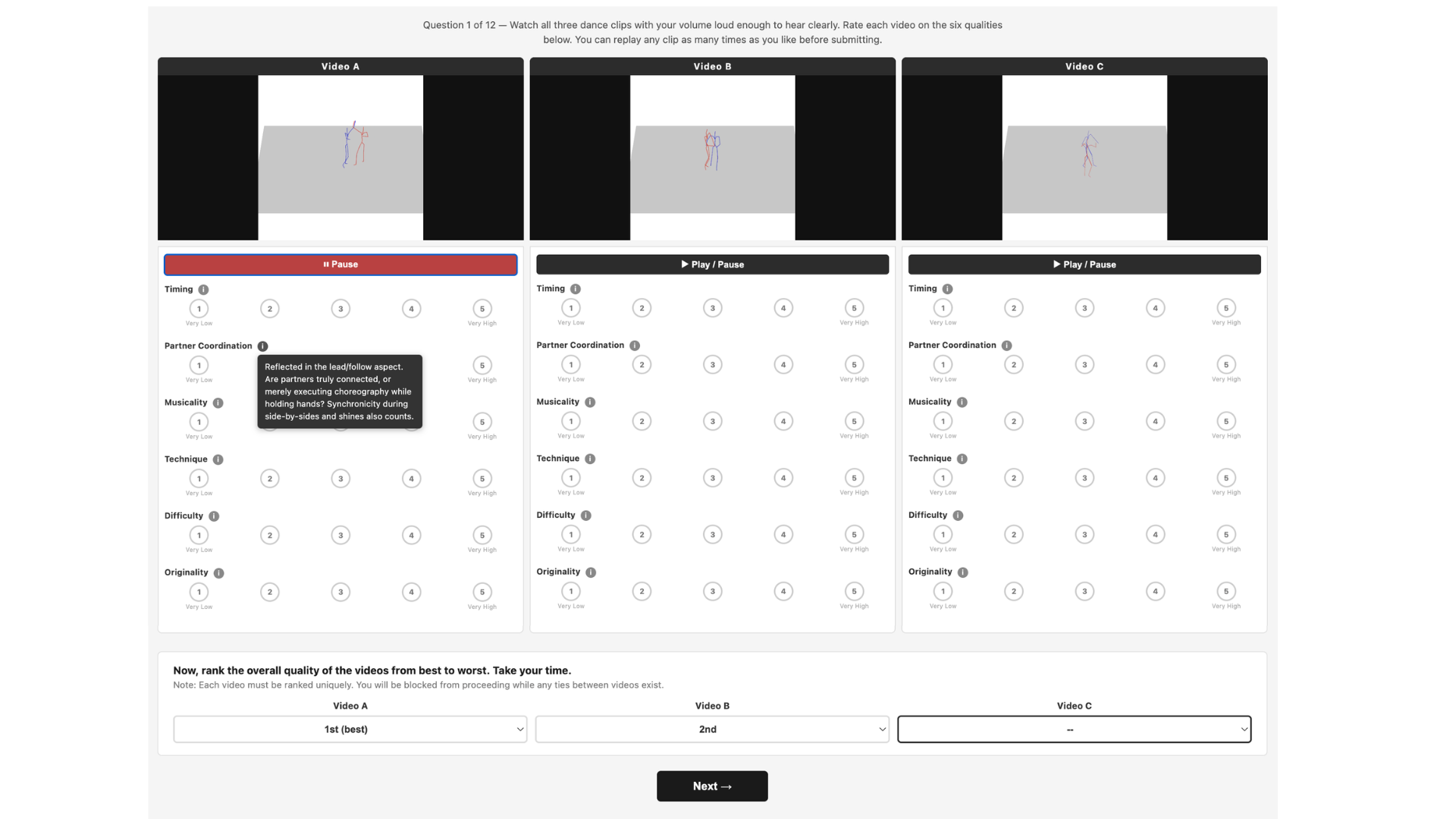}
    \caption{The human evaluation interface presented to Prolific participants. Participants rated each video on six dimensions using 5-point Likert scales.}
    \label{Fig:study_screenshots}
\end{figure*}

For every video stimulus, participants rated each clip along six dimensions of dance quality using a 5-point Likert scale. After completing the per-video ratings, participants ranked the videos overall from best to worst. Figure~\ref{Fig:study_screenshots} presents this evaluation interface. Participants viewed 12 consecutive trials, each comparing 3 videos along the six dimensions described below. Overall video rankings were collected at the end of each comparison trial, as shown in Figure~\ref{Fig:study_screenshots}. 

Participants accessed dimension definitions by hovering their cursor over any \textbf{i} icon in the interface. These definitions are derived from partnered salsa competition evaluation criteria and are presented in Table~\ref{tab:dimension_definitions}. 

\begin{table*}[h]
\centering
\small
\renewcommand{\arraystretch}{1.5}
\begin{tabular}{|l|p{9cm}|}
\hline
\textbf{Dimension} & \textbf{Definition shown to participants} \\
\hline
Timing &
    The majority of the routine should show recognizable timing in the dance genre (i.e.\ Salsa: 1-2-3, 5-6-7). \\
\hline
Partner Coordination &
    Reflected in the lead/follow aspect. Are partners truly connected, or merely executing choreography while holding hands? Synchronicity during side-by-sides and shines also counts. \\
\hline
Musicality &
    The couple's ability to creatively work with the timing of the music, including through patterns, footwork, hits, tricks, etc. \\
\hline
Technique &
    Reflected through balance, placement and line; sometimes called latin motion or sabor. Movement should be grounded, with clarity of weight changes, and appear both clear and effortless. \\
\hline
Difficulty &
    Reflected by difficulty of patterns, turns, shines, and level of tricks, dips, drops and flares. Movement must be successfully executed to receive credit. \\
\hline
Originality &
    The couple's interpretation of the music. Includes clever turn patterns, shines, musical choreography, and good use of space. \\
\hline
\end{tabular}
\caption{Evaluation dimensions and their definitions, as displayed to participants in the human study interface. Adapted from Salsa Congress Rules\cite{canadasalsacongress_rules_2024}. Compared to the competition rules, only "showmanship", which rates costumes and good sportsmanship, was omitted. The competition item entitled "choreography" was renamed to originality (due to the improvised nature of the dataset), but the definition remains similar.}
\label{tab:dimension_definitions}
\end{table*}

\section{Classification Details}

\begin{figure*}[t]
    \centering
    \begin{subfigure}[t]{0.48\linewidth}
        \centering
        \includegraphics[width=\linewidth]{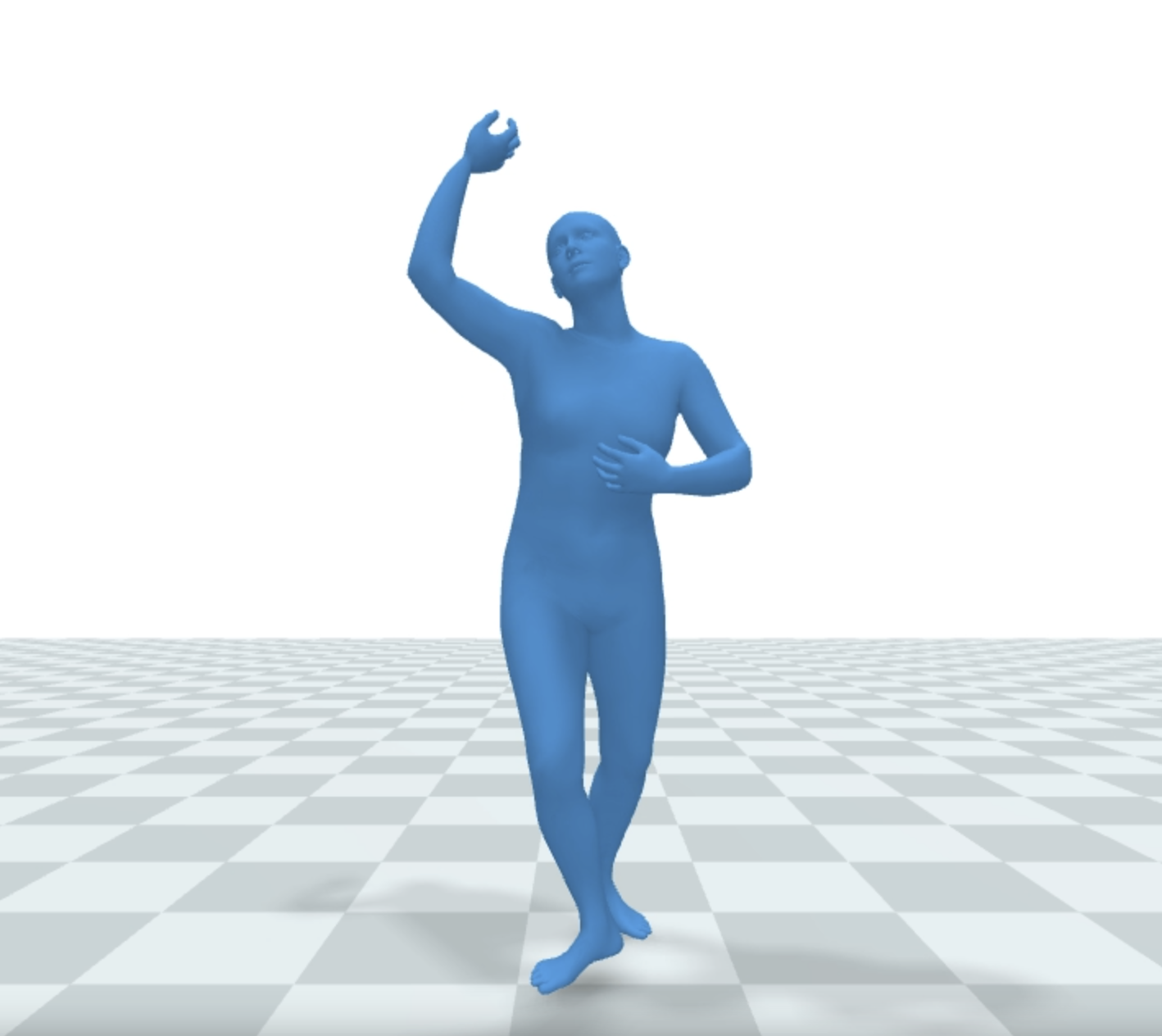}
        
    \end{subfigure}
    \hfill
    \begin{subfigure}[t]{0.48\linewidth}
        \centering
        \includegraphics[width=\linewidth]{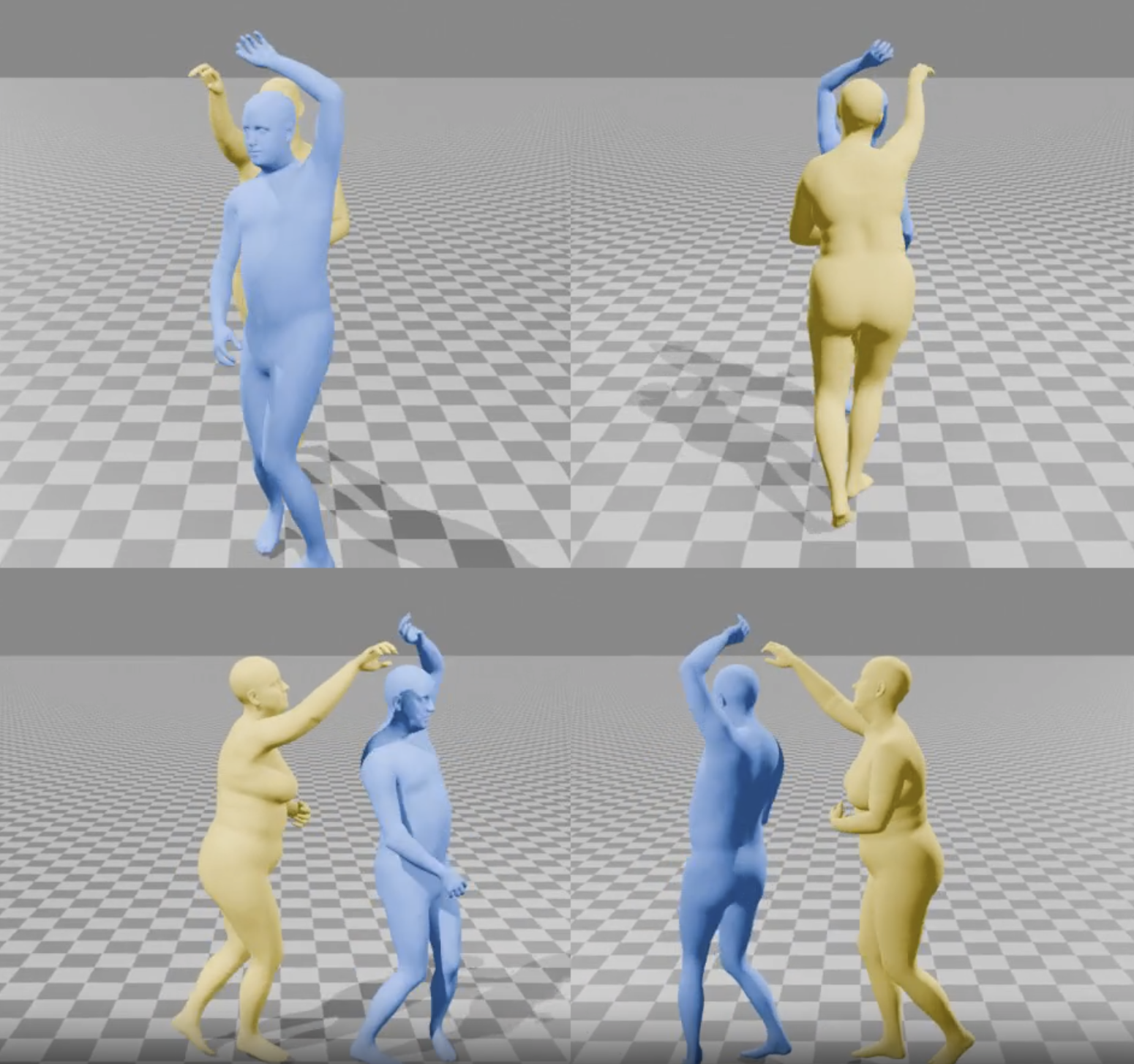}
    \end{subfigure}
    \vspace{0.5em}
    \caption{Sample frames from the two video configurations used in our experiments: (1) follower-only (single person) (\textit{left}) and (2) follower and leader (dyadic) (\textit{right}).}
    \label{fig:single-duet}
\end{figure*}


We performed both move classification and proficiency estimation using two approaches: (1) fine-tuning vision-language models (VLMs), specifically Qwen2.5-VL~\footnote{https://huggingface.co/Qwen/Qwen2.5-VL-7B-Instruct}, LLaVA-NeXT-Video~\footnote{https://huggingface.co/llava-hf/LLaVA-NeXT-Video-7B-hf}, and InternVL3~\footnote{https://huggingface.co/OpenGVLab/InternVL3-8B-hf}, and (2) evaluating them in a zero-shot setting. Both approaches were applied to two video configurations: follower-only videos and videos containing both the follower and the leader (Fig~\ref{fig:single-duet}). Note that for videos containing both the leader and the follower, we use multiple camera viewpoints to ensure that the movements of each dancer are fully captured.

The experiments were conducted on a single NVIDIA GeForce RTX 3090 GPU.
For fine-tuning we utilized LoRA~\cite{hu2022lora} with 4-bit quantization~\cite{dettmers2023qlora}. We used a rank of 16, \texttt{lora\_alpha} of 32, and applied LoRA to the following target modules: \texttt{q\_proj}, \texttt{k\_proj}, \texttt{v\_proj}, \texttt{o\_proj}, \texttt{gate\_proj}, \texttt{up\_proj}, and \texttt{down\_proj}. We optimized with \texttt{paged\_adamw\_8bit} at a learning rate of $2 \times 10^{-4}$.

Models were trained for 3 epochs, and we report results from the epoch with the best performance on the test set. This selection is motivated by our intention to reuse the fine-tuned model as an evaluation tool on generated data.

For each video segment, we uniformly sample 8 frames as input to the VLM. To manage computational cost, we resize frames such that the shorter side is at most 512 pixels, preserving the original aspect ratio.

During inference, we set \texttt{do\_sample=False} and \texttt{max\_new\_tokens=16}. Rather than requiring an exact string match, we accept a prediction if exactly one valid label appears anywhere in the generated text. For example, ``-XBL'' is accepted, while ``XBL, Copa'' is rejected due to the presence of multiple labels.




\subsection{Move Classification}
\begin{figure*}[t]
    \centering
    \begin{subfigure}[t]{0.48\linewidth}
        \centering
        \includegraphics[width=\linewidth]{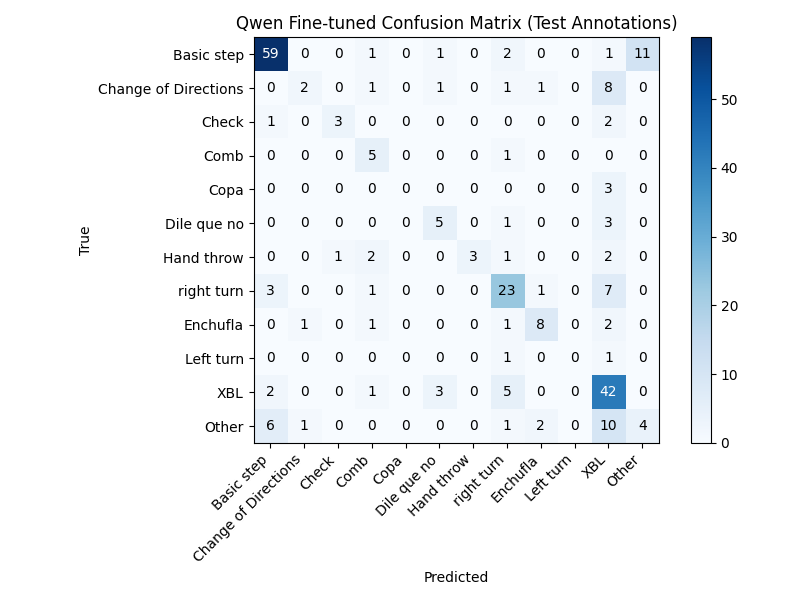}
        \caption{Fine-tuned Qwen2.5-VL on follower only videos.}
        \label{fig:ft_q_single_move}
    \end{subfigure}
    \hfill
    \begin{subfigure}[t]{0.48\linewidth}
        \centering
        \includegraphics[width=\linewidth]{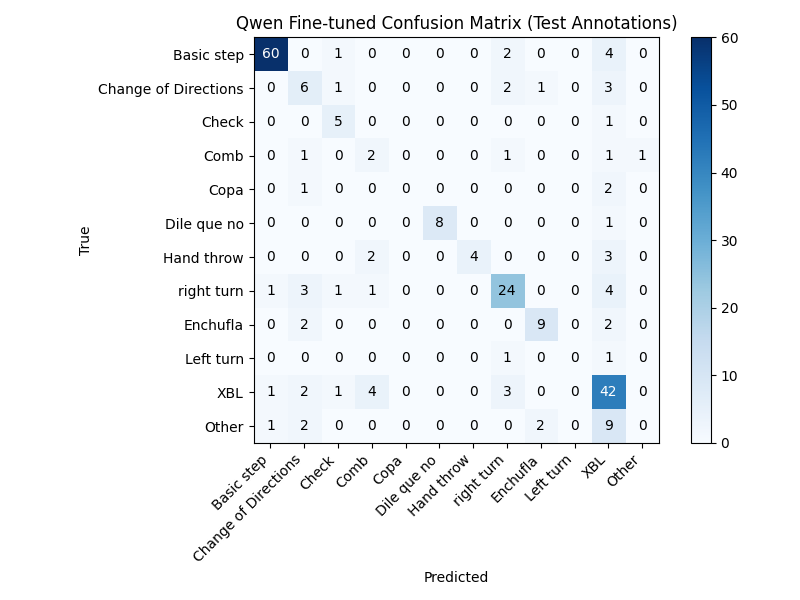}
        \caption{Fine-tuned Qwen2.5-VL on follower \& leader videos.}
        \label{fig:ft_q_duet_move}
    \end{subfigure}
    \vspace{0.5em}
    \begin{subfigure}[t]{0.48\linewidth}
        \centering
        \includegraphics[width=\linewidth]{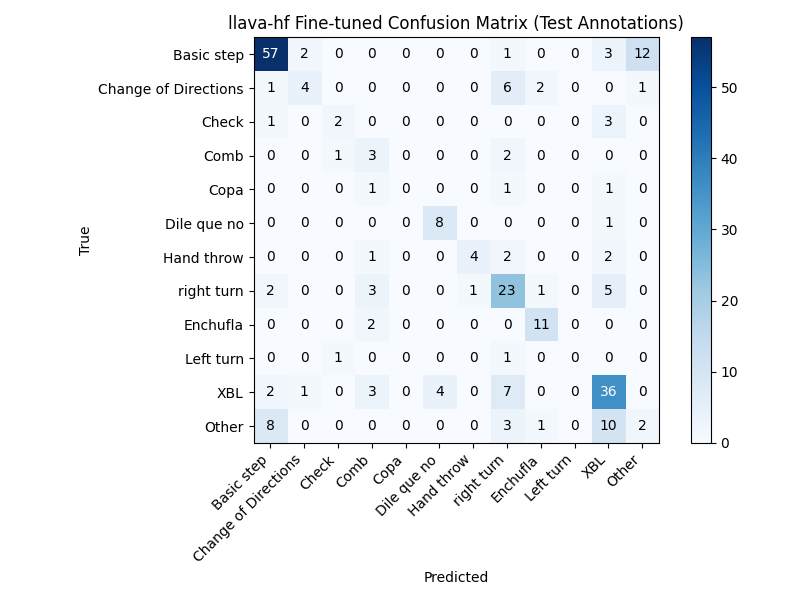}
        \caption{Fine-tuned LLaVA on follower only videos.}
        \label{fig:ft_llava_single_move}
    \end{subfigure}
    \hfill
    \begin{subfigure}[t]{0.48\linewidth}
        \centering
        \includegraphics[width=\linewidth]{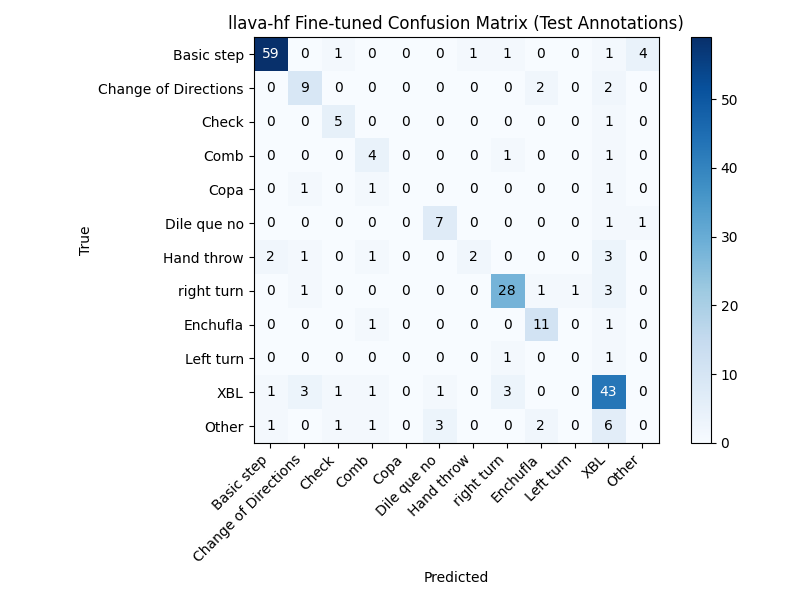}
        \caption{Fine-tuned LLaVA on follower \& leader videos.}
        \label{fig:ft_llava_duet_move}
    \end{subfigure}
    \vspace{0.5em}
    \begin{subfigure}[t]{0.48\linewidth}
        \centering
        \includegraphics[width=\linewidth]{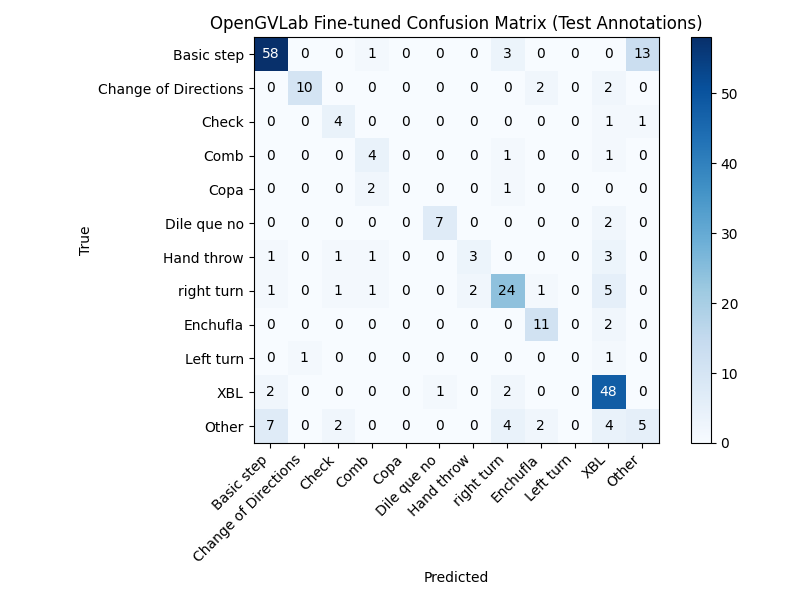}
        \caption{Fine-tuned InternVL3 on follower only videos.}
        \label{fig:ft_intern_single_move}
    \end{subfigure}
    \hfill
    \begin{subfigure}[t]{0.48\linewidth}
        \centering
        \includegraphics[width=\linewidth]{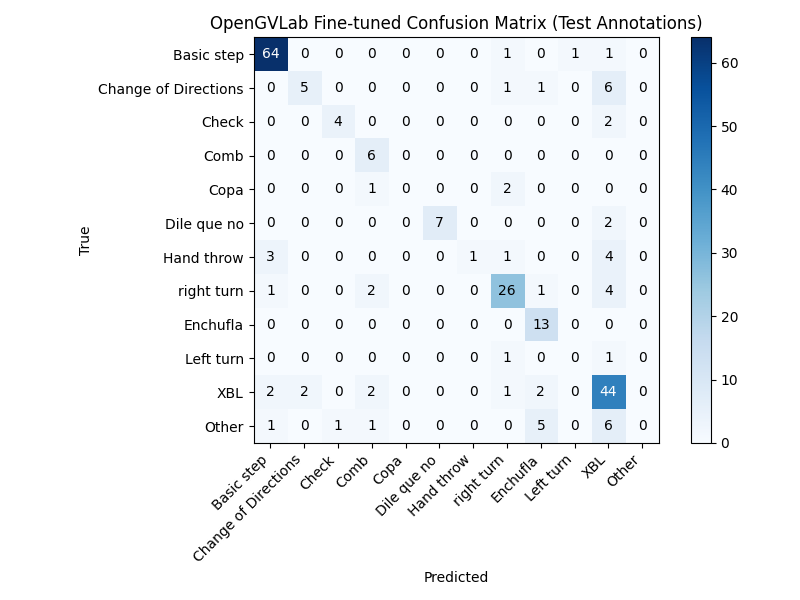}
        \caption{Fine-tuned InternVL3 on follower \& leader videos.}
        \label{fig:ft_intern_duet_move}
    \end{subfigure}
    \caption{Confusion matrices for move classification using fine-tuned models, reported in Table~\ref{tab:classification}.}
    \label{fig:cm_finetuned_move}
\end{figure*}

\begin{figure*}[t]
    \centering
    \begin{subfigure}[t]{0.48\linewidth}
        \centering
        \includegraphics[width=\linewidth]{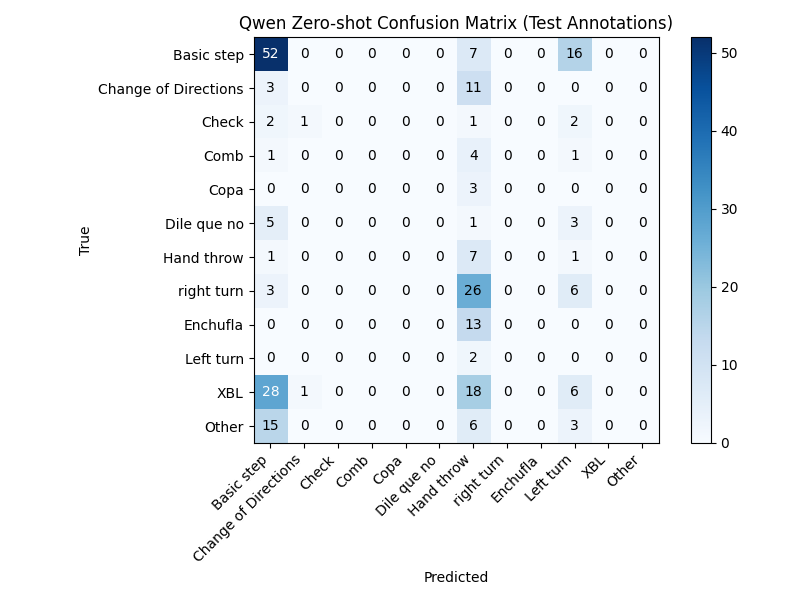}
        \caption{Zero-shot Qwen2.5-VL on follower only videos.}
        \label{fig:zs_q_single_move}
    \end{subfigure}
    \hfill
    \begin{subfigure}[t]{0.48\linewidth}
        \centering
        \includegraphics[width=\linewidth]{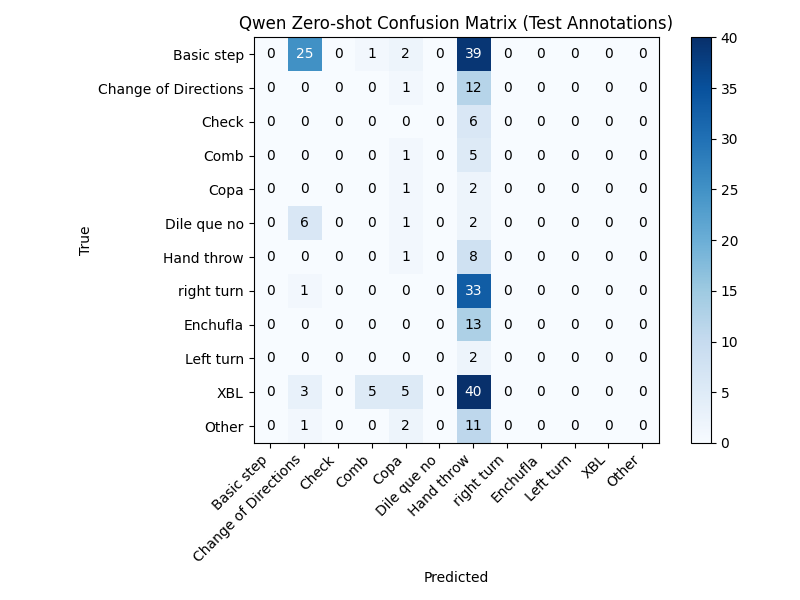}
        \caption{Zero-shot Qwen2.5-VL on follower \& leader videos.}
        \label{fig:zs_q_duet_move}
    \end{subfigure}
    \vspace{0.5em}
    \begin{subfigure}[t]{0.48\linewidth}
        \centering
        \includegraphics[width=\linewidth]{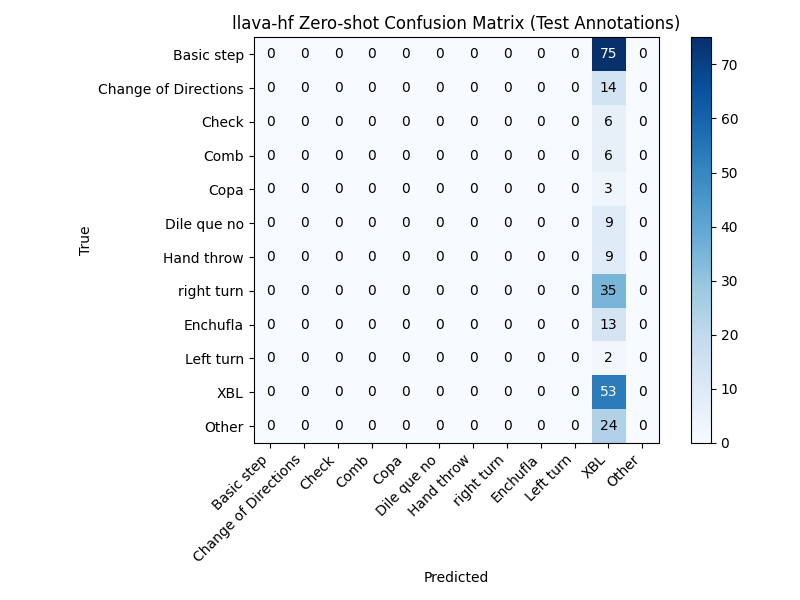}
        \caption{Zero-shot LLaVA on follower only videos.}
        \label{fig:zs_llava_single_move}
    \end{subfigure}
    \hfill
    \begin{subfigure}[t]{0.48\linewidth}
        \centering
        \includegraphics[width=\linewidth]{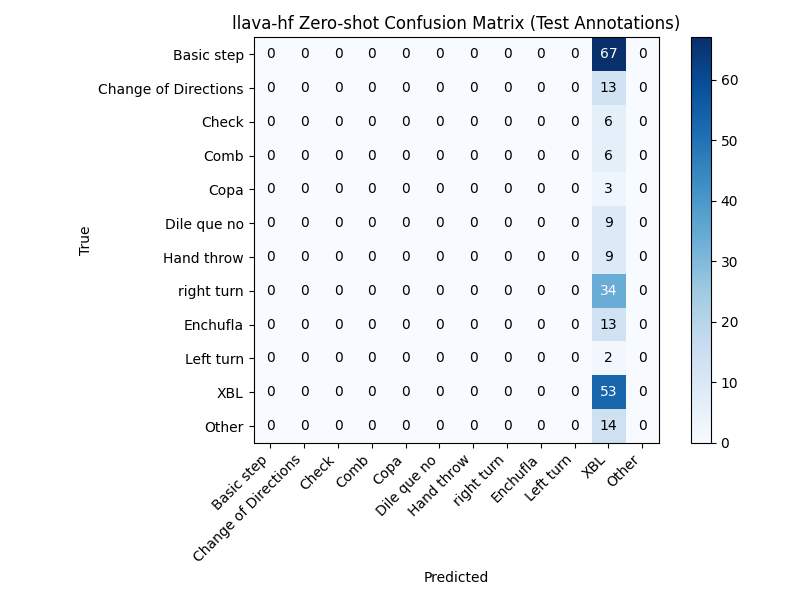}
        \caption{Zero-shot LLaVA on follower \& leader videos.}
        \label{fig:zs_llava_duet_move}
    \end{subfigure}
    \vspace{0.5em}
    \begin{subfigure}[t]{0.48\linewidth}
        \centering
        \includegraphics[width=\linewidth]{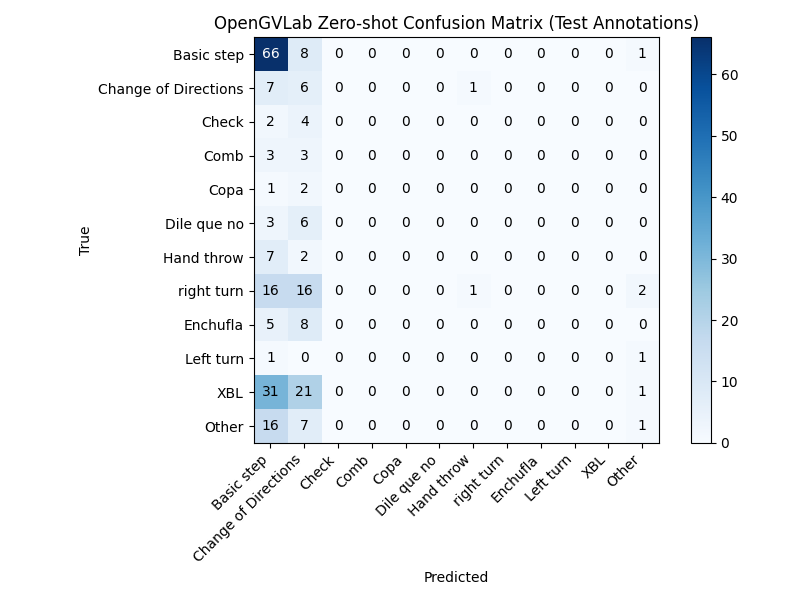}
        \caption{Zero-shot InternVL3 on follower only videos.}
        \label{fig:zs_intern_single_move}
    \end{subfigure}
    \hfill
    \begin{subfigure}[t]{0.48\linewidth}
        \centering
        \includegraphics[width=\linewidth]{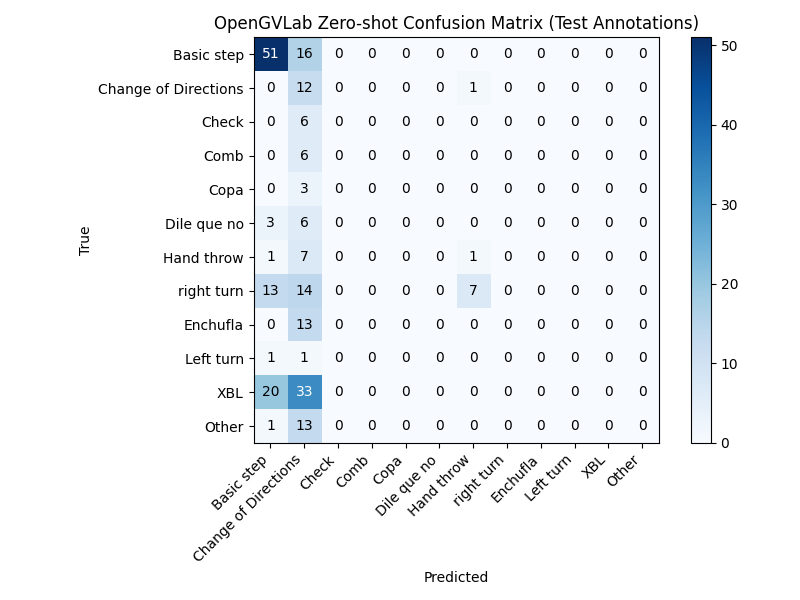}
        \caption{Zero-shot InternVL3 on follower \& leader videos.}
        \label{fig:zs_intern_duet_move}
    \end{subfigure}
    \caption{Confusion matrices for move classification using zero-shot models, reported in Table~\ref{tab:classification}.}
    \label{fig:cm_zeroshot_move}
\end{figure*}

We derive our label set from the annotated videos by retaining only moves that appear at least 20 times in the training set and are jointly performed by both the leader and the follower. This yields 11 move classes: \textit{Basic Step}, \textit{Change of Direction}, \textit{Check}, \textit{Comb}, \textit{Copa}, \textit{Dile que no}, \textit{Hand Throw}, \textit{Right Turn}, \textit{Enchufla}, \textit{Left Turn}, and \textit{XBL}. Any annotation whose base move does not appear in this list is assigned to an additional \textit{Other} class.

The annotation subset used for training and evaluation depends on the video configuration. For videos containing both the leader and the follower, we use only annotations where the role is \textit{Leader\&Follower}. For follower-only videos, we include annotations where the role is either \textit{Leader\&Follower} or \textit{Follower}, as the follower is always visible regardless of configuration.

We use the following prompt:
\begin{quote}
\textit{You are a dance move classifier.\\
Look at this video clip and pick exactly ONE best label from the list below.\\
Return only the label text exactly as written, with no explanation.\\ \\
Candidate labels:\\
\textless List of all possible labels\textgreater}
\end{quote}

Fig.~\ref{fig:cm_finetuned_move} and Fig.~\ref{fig:cm_zeroshot_move} show the confusion matrices for move classification using fine-tuned and zero-shot models, respectively, corresponding to the results reported in Table~\ref{tab:classification}. While zero-shot models tend to over-predict a single label, fine-tuned models produce more balanced predictions across all classes, resulting in substantially higher F1 scores.

\subsection{Proficiency Estimation}
\label{appendix-prof-est}
\begin{figure*}[t]
    \centering
    \begin{subfigure}[t]{0.48\linewidth}
        \centering
        \includegraphics[width=\linewidth]{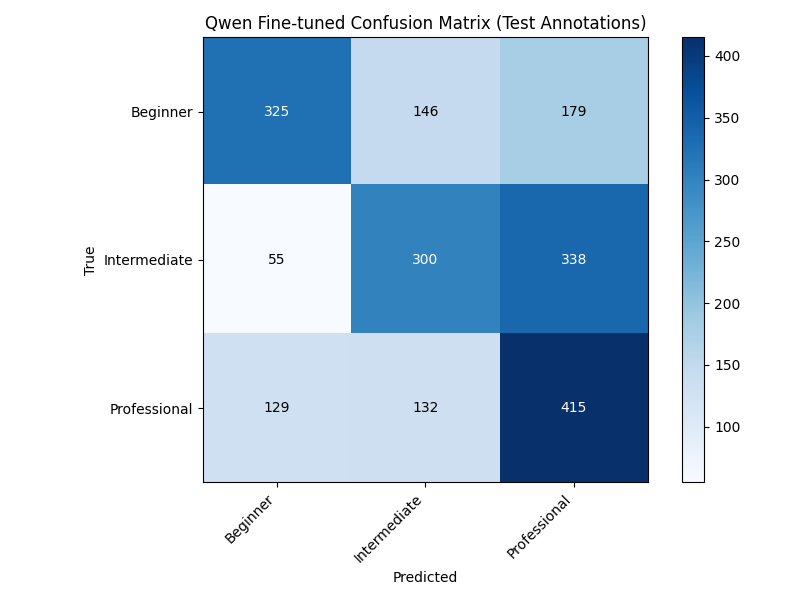}
        \caption{Fine-tuned Qwen2.5-VL on follower only videos.}
        \label{fig:ft_q_single}
    \end{subfigure}
    \hfill
    \begin{subfigure}[t]{0.48\linewidth}
        \centering
        \includegraphics[width=\linewidth]{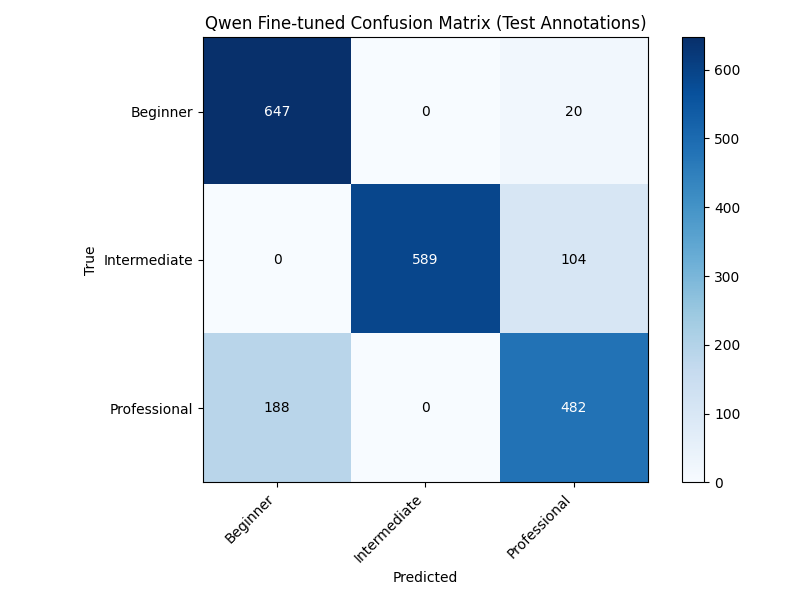}
        \caption{Fine-tuned Qwen2.5-VL on follower \& leader videos.}
        \label{fig:ft_q_duet}
    \end{subfigure}
    \vspace{0.5em}
    \begin{subfigure}[t]{0.48\linewidth}
        \centering
        \includegraphics[width=\linewidth]{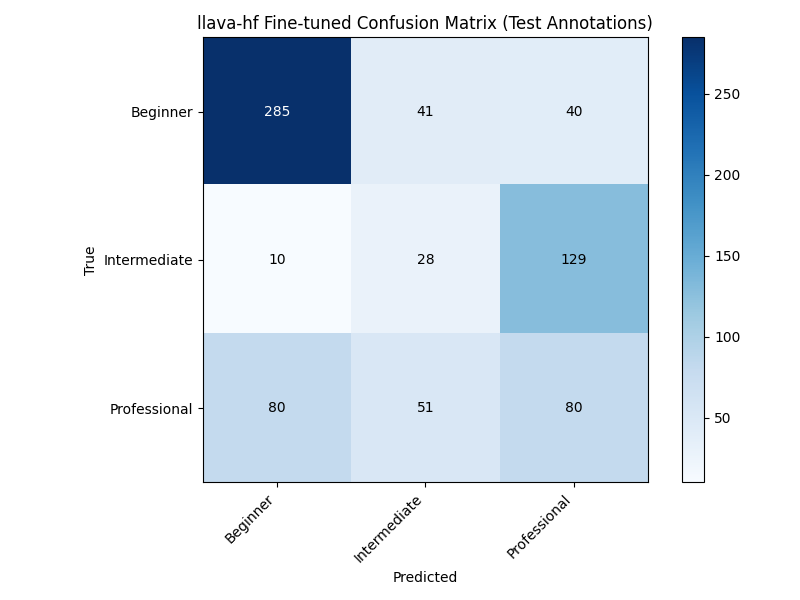}
        \caption{Fine-tuned LLaVA on follower only videos.}
        \label{fig:ft_llava_single}
    \end{subfigure}
    \hfill
    \begin{subfigure}[t]{0.48\linewidth}
        \centering
        \includegraphics[width=\linewidth]{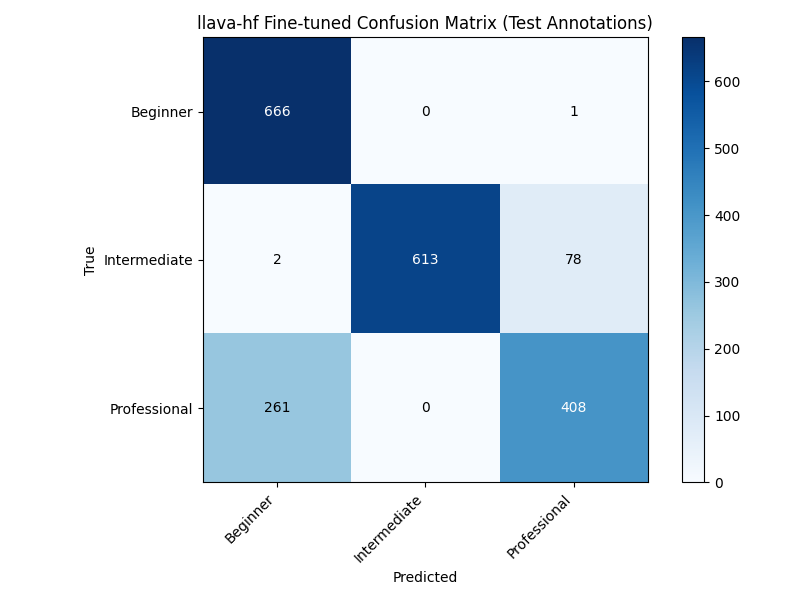}
        \caption{Fine-tuned LLaVA on follower \& leader videos.}
        \label{fig:ft_llava_duet}
    \end{subfigure}
    \vspace{0.5em}
    \begin{subfigure}[t]{0.48\linewidth}
        \centering
        \includegraphics[width=\linewidth]{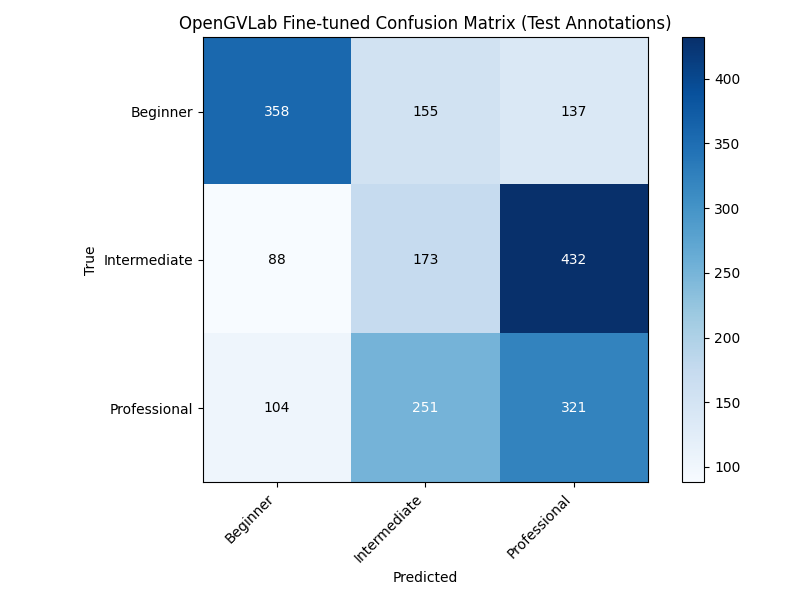}
        \caption{Fine-tuned InternVL3 on follower only videos.}
        \label{fig:ft_intern_single}
    \end{subfigure}
    \hfill
    \begin{subfigure}[t]{0.48\linewidth}
        \centering
        \includegraphics[width=\linewidth]{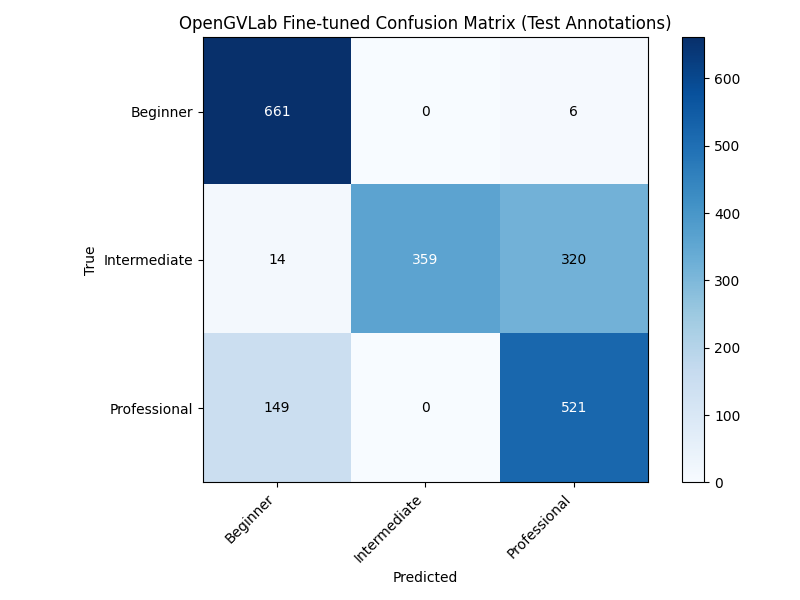}
        \caption{Fine-tuned InternVL3 on follower \& leader videos.}
        \label{fig:ft_intern_duet}
    \end{subfigure}
    \caption{Confusion matrices for proficiency classification using fine-tuned models, reported in Table~\ref{tab:classification}.}
    \label{fig:cm_finetuned_level}
\end{figure*}

\begin{figure*}[t]
    \centering
    \begin{subfigure}[t]{0.48\linewidth}
        \centering
        \includegraphics[width=\linewidth]{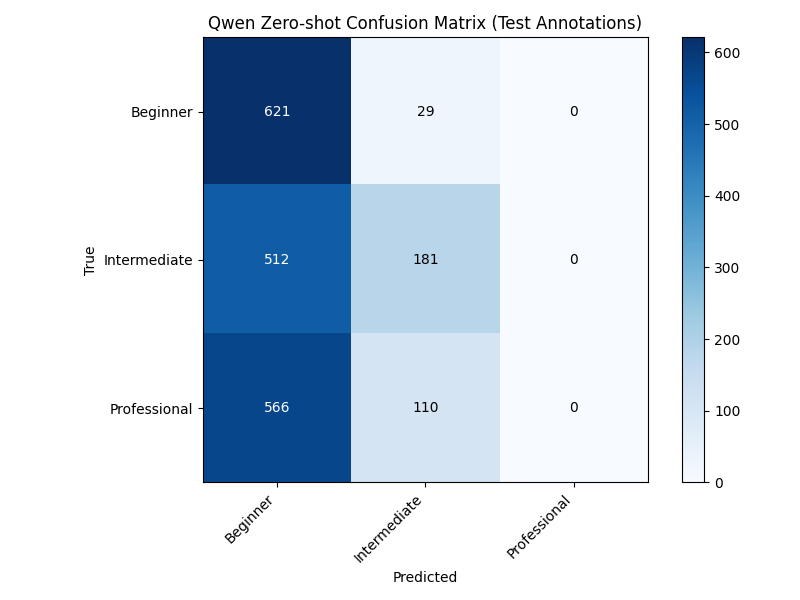}
        \caption{Zero-shot Qwen2.5-VL on follower only videos.}
        \label{fig:zs_q_single}
    \end{subfigure}
    \hfill
    \begin{subfigure}[t]{0.48\linewidth}
        \centering
        \includegraphics[width=\linewidth]{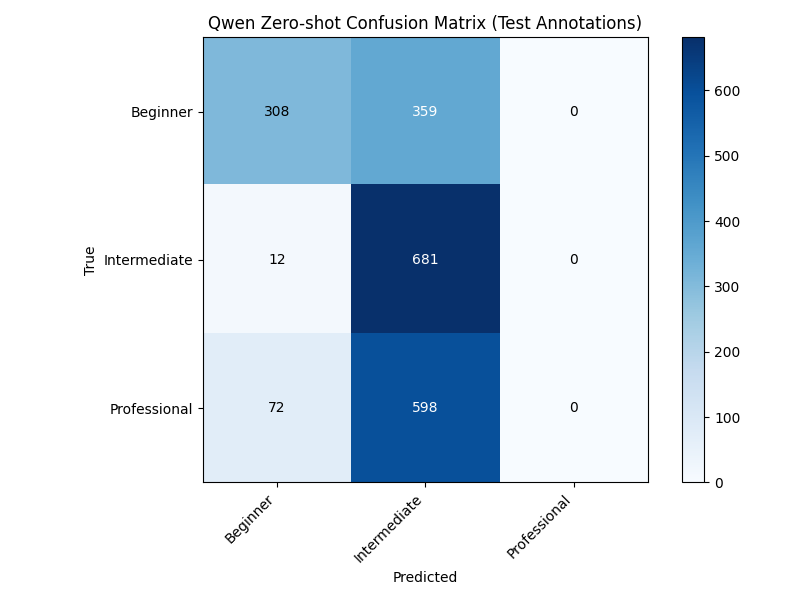}
        \caption{Zero-shot Qwen2.5-VL on follower \& leader videos.}
        \label{fig:zs_q_duet}
    \end{subfigure}
    \vspace{0.5em}
    \begin{subfigure}[t]{0.48\linewidth}
        \centering
        \includegraphics[width=\linewidth]{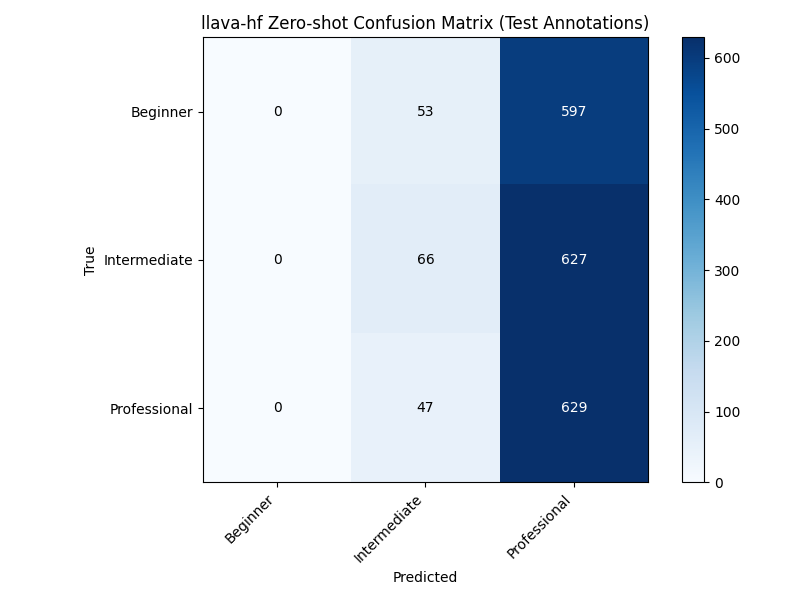}
        \caption{Zero-shot LLaVA on follower only videos.}
        \label{fig:zs_llava_single}
    \end{subfigure}
    \hfill
    \begin{subfigure}[t]{0.48\linewidth}
        \centering
        \includegraphics[width=\linewidth]{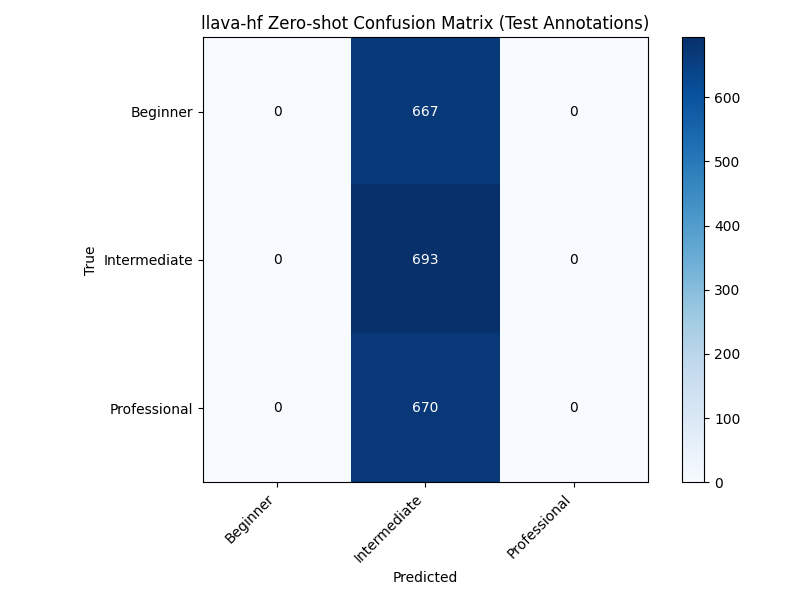}
        \caption{Zero-shot LLaVA on follower \& leader videos.}
        \label{fig:zs_llava_duet}
    \end{subfigure}
    \vspace{0.5em}
    \begin{subfigure}[t]{0.48\linewidth}
        \centering
        \includegraphics[width=\linewidth]{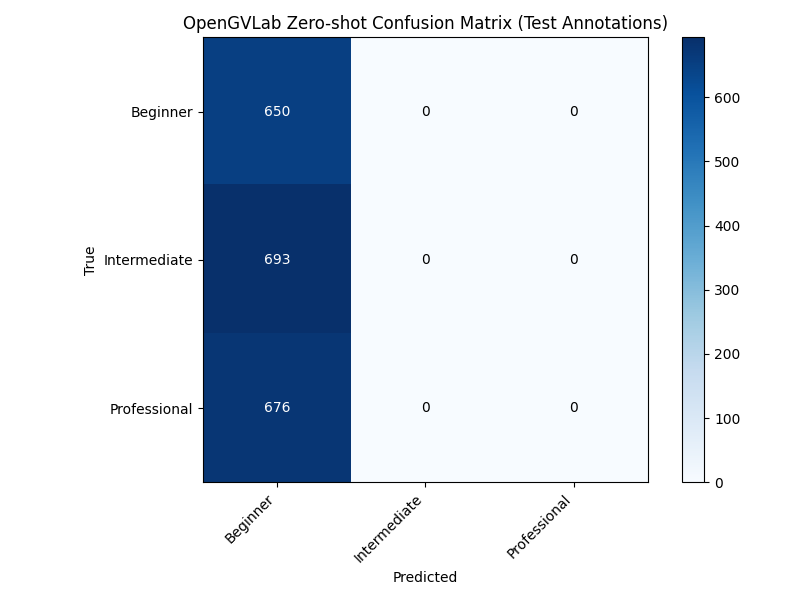}
        \caption{Zero-shot InternVL3 on follower only videos.}
        \label{fig:zs_intern_single}
    \end{subfigure}
    \hfill
    \begin{subfigure}[t]{0.48\linewidth}
        \centering
        \includegraphics[width=\linewidth]{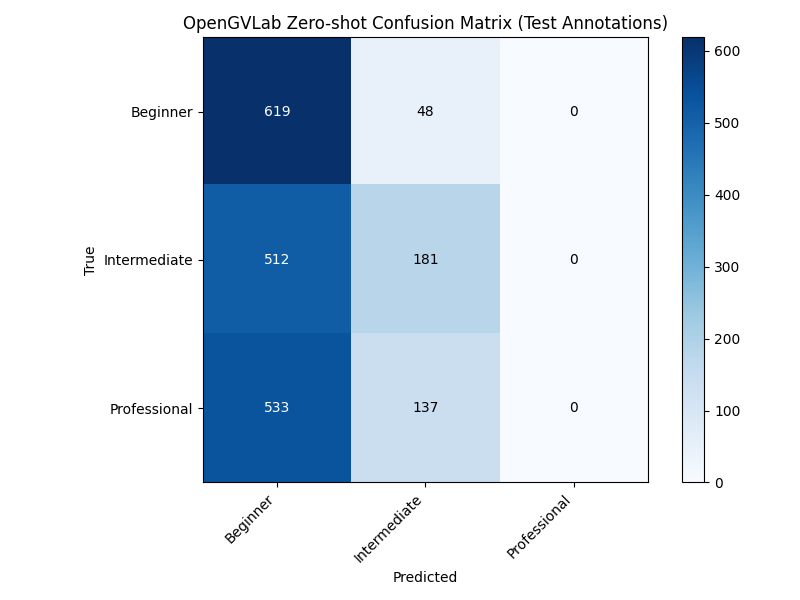}
        \caption{Zero-shot InternVL3 on follower \& leader videos.}
        \label{fig:zs_intern_duet}
    \end{subfigure}
    \caption{Confusion matrices for proficiency classification using zero-shot models, reported in Table~\ref{tab:classification}.}
    \label{fig:cm_zeroshot_level}
\end{figure*}

We define three proficiency levels: Beginner (Pair1, Pair3, Pair8), Intermediate (Pair2, Pair4, Pair6), and Professional (Pair5, Pair7, Pair9). To prevent the model from learning to associate specific pair-specific movements with their proficiency level, we held out three pairs (one pair per level: Pair6, Pair8, and Pair9) exclusively for testing (not used for training).

To construct samples for level classification, we segment each pair's video into clips of varying durations, sampled randomly between the minimum and maximum duration of annotated moves. 

We use the following prompt:
\begin{quote}
\textit{You are a dance proficiency level classifier.\\
Look at this video clip and pick exactly ONE best label from the list below.\\
Return only the label text exactly as written, with no explanation.\\ \\
Candidate labels:\\
\textless List of all possible labels\textgreater}
\end{quote}
  
Fig.~\ref{fig:cm_finetuned_level} and Fig.~\ref{fig:cm_zeroshot_level} show the confusion matrices for proficiency estimation using fine-tuned and zero-shot models, respectively, corresponding to the results reported in Table~\ref{tab:classification}. Similar to move classification, zero-shot models tend to over-predict a single label, whereas fine-tuned models produce more balanced predictions across all classes.





\subsection{Objective Metrics}

\begin{figure*}[t]
    \centering
    \begin{subfigure}[t]{0.48\linewidth}
        \centering
        \includegraphics[width=\linewidth]{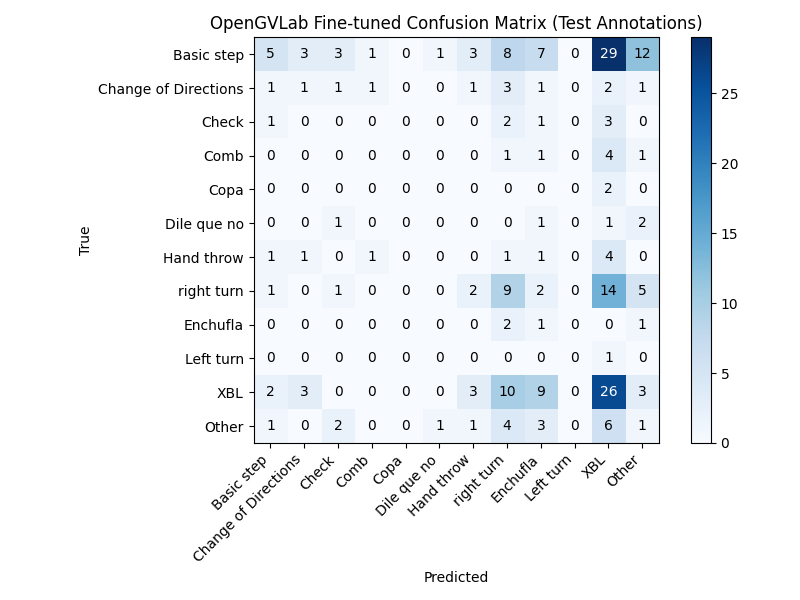}
        \caption{InterGen with fine-tuned InternVL3.}
    \end{subfigure}
    \hfill
    \begin{subfigure}[t]{0.48\linewidth}
        \centering
        \includegraphics[width=\linewidth]{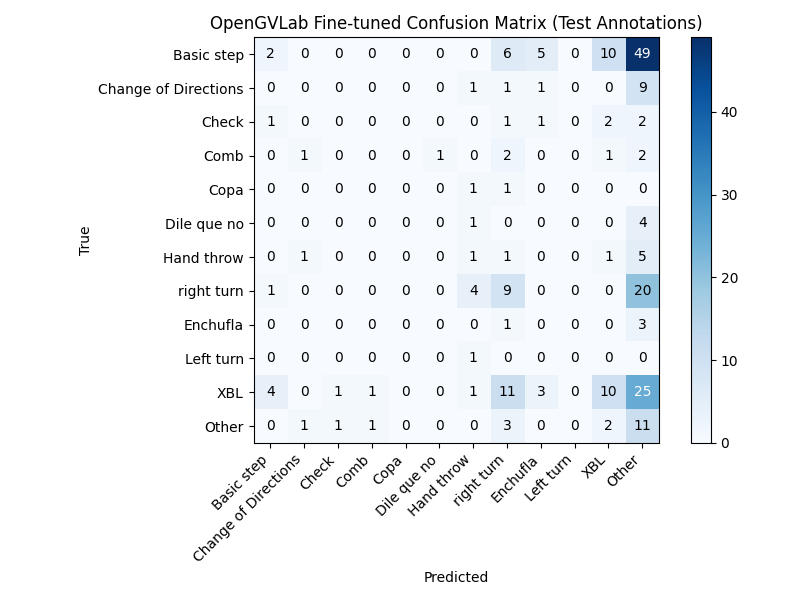}
        \caption{Duolando with fine-tuned InternVL3.}
    \end{subfigure}
    \vspace{0.5em}
    \begin{subfigure}[t]{0.48\linewidth}
        \centering
        \includegraphics[width=\linewidth]{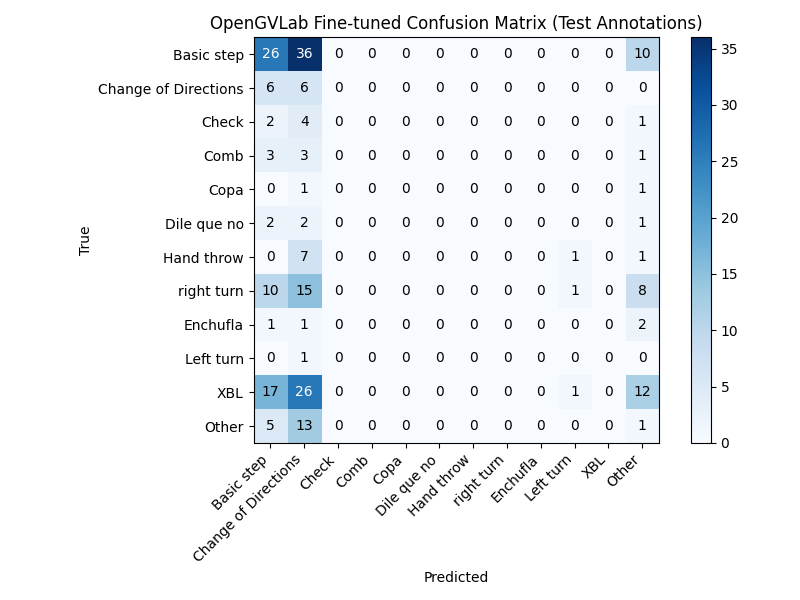}
        \caption{InterGen with zero-shot InternVL3.}
    \end{subfigure}
    \hfill
    \begin{subfigure}[t]{0.48\linewidth}
        \centering
        \includegraphics[width=\linewidth]{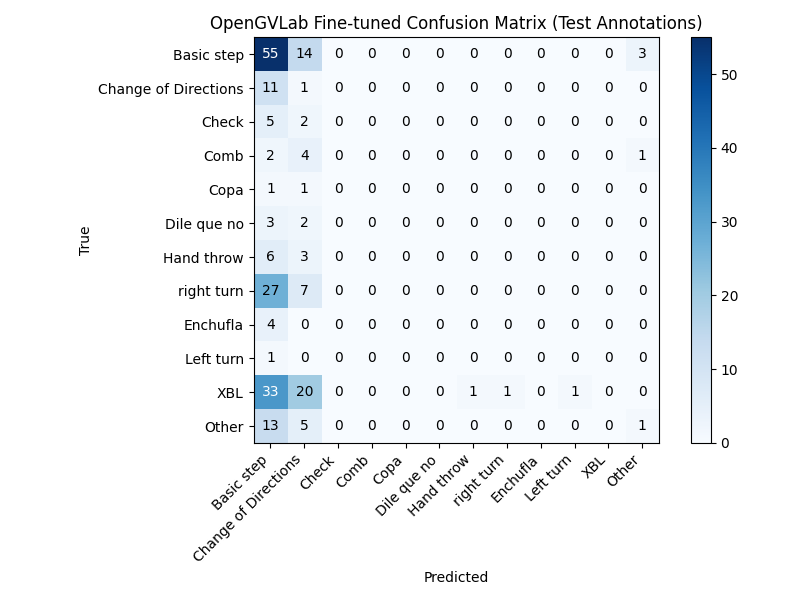}
        \caption{Duolando with zero-shot InternVL3.}
    \end{subfigure}
    \vspace{0.5em}

    \caption{Confusion matrices for legibility using both zero-shot and fine-tuned InternVL3 models, reported in Table~\ref{tab:contextual}.}
    \label{fig:cm_legibility_move}
\end{figure*}

\begin{figure*}[t]
    \centering
    \begin{subfigure}[t]{0.48\linewidth}
        \centering
        \includegraphics[width=\linewidth]{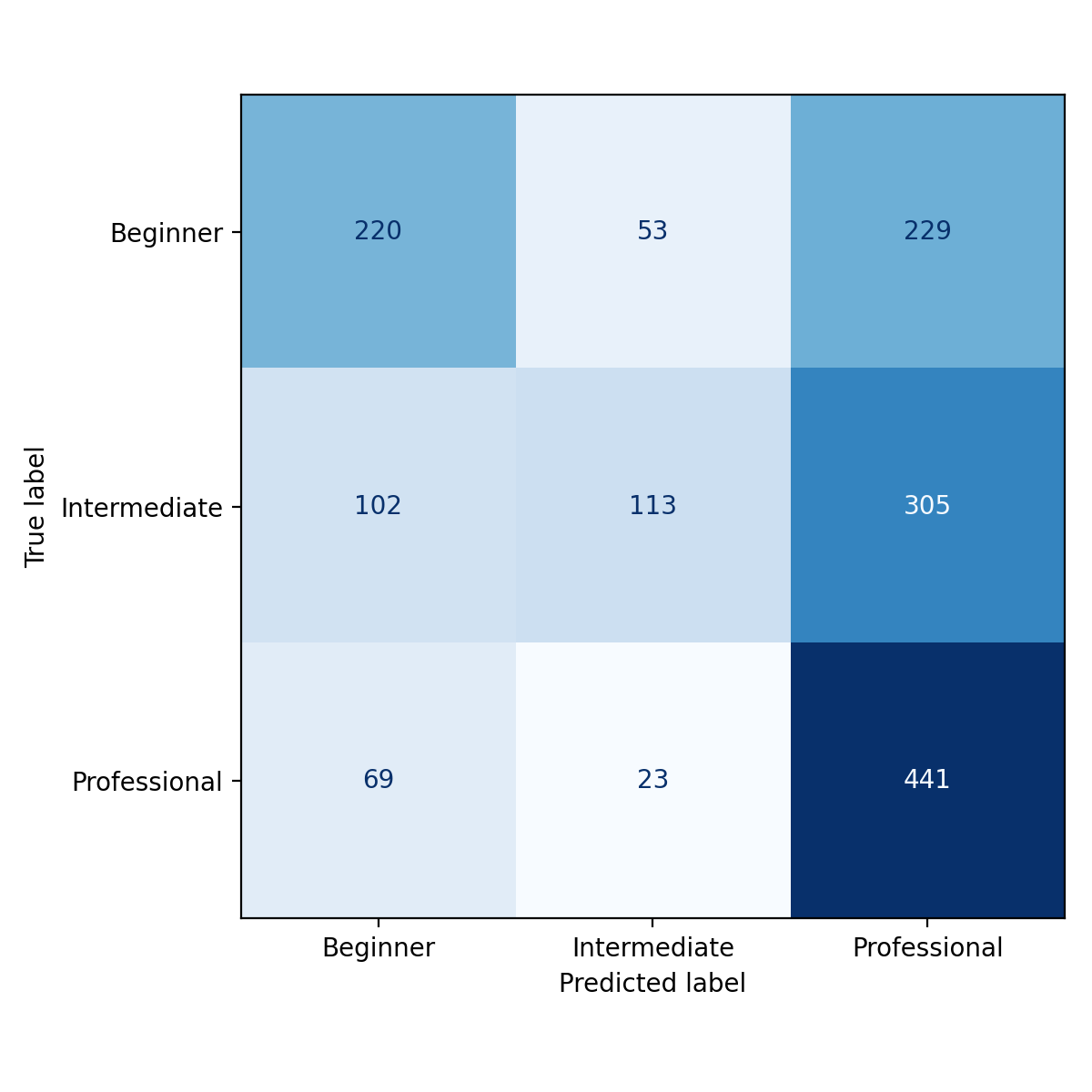}
        \caption{InterGen with fine-tuned Qwen2.5-VL.}
    \end{subfigure}
    \hfill
    \begin{subfigure}[t]{0.48\linewidth}
        \centering
        \includegraphics[width=\linewidth]{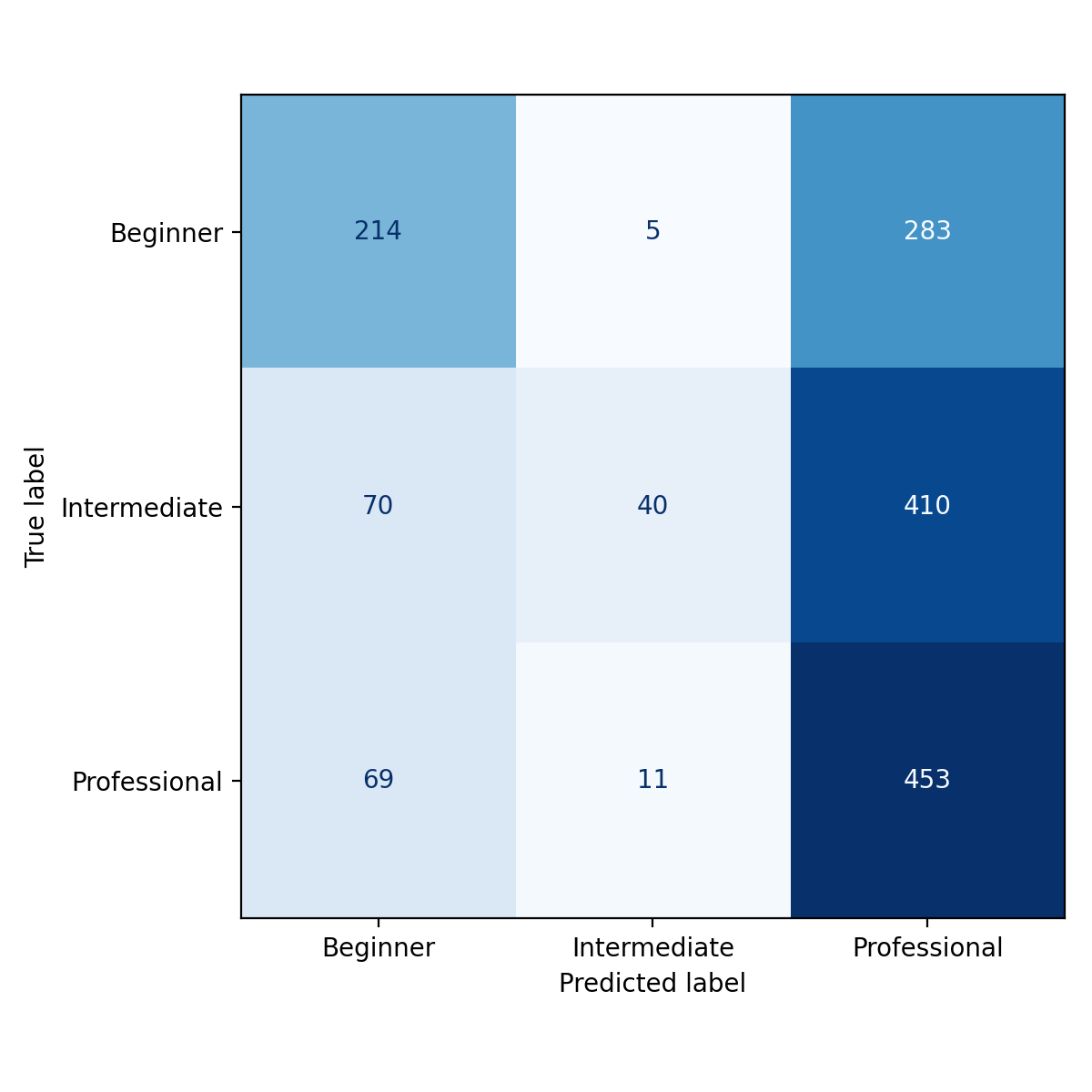}
        \caption{Dualando with fine-tuned Qwen2.5-VL.}
    \end{subfigure}
    \vspace{0.5em}
    \begin{subfigure}[t]{0.48\linewidth}
        \centering
        \includegraphics[width=\linewidth]{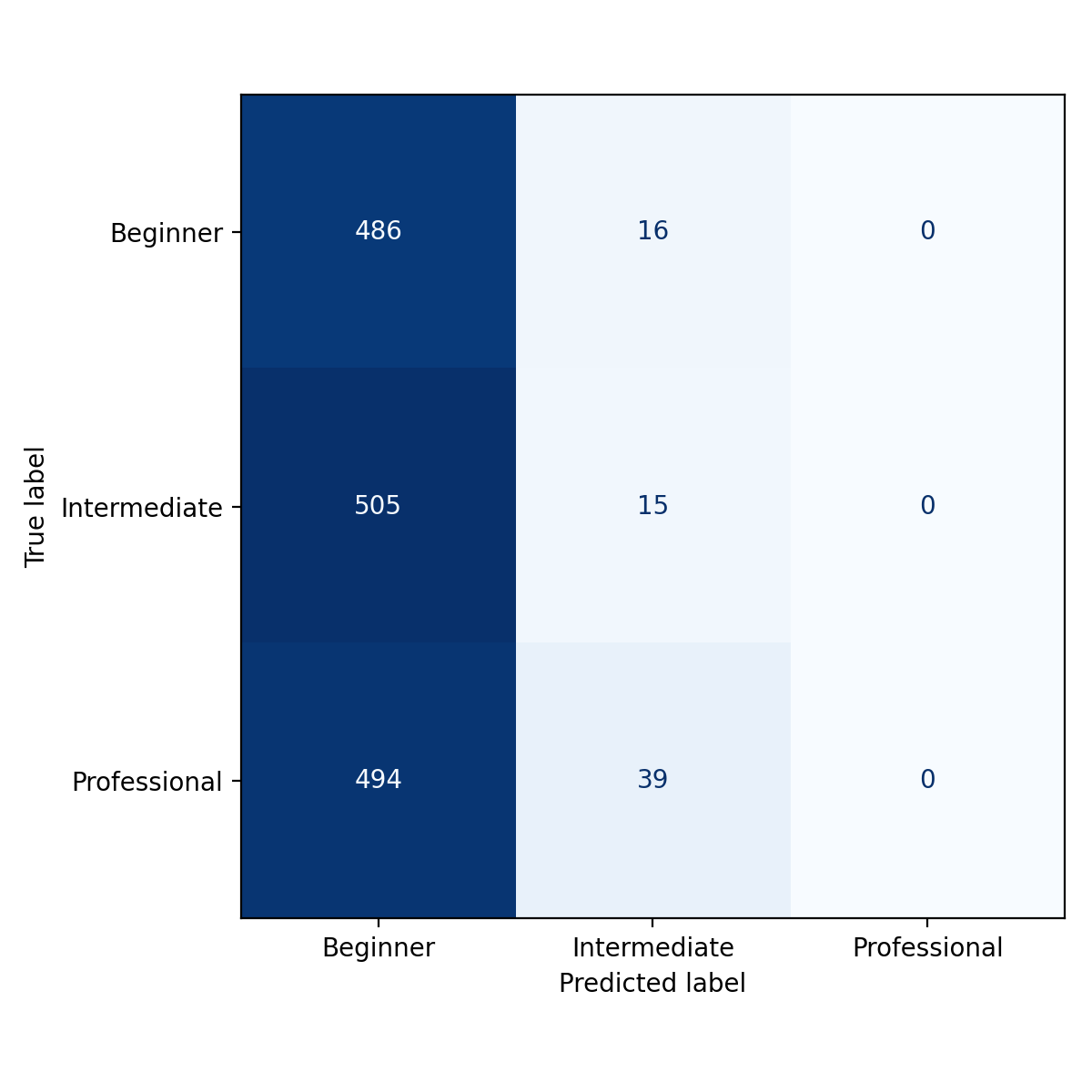}
        \caption{InterGen with zero-shot Qwen2.5-VL.}
    \end{subfigure}
    \hfill
    \begin{subfigure}[t]{0.48\linewidth}
        \centering
        \includegraphics[width=\linewidth]{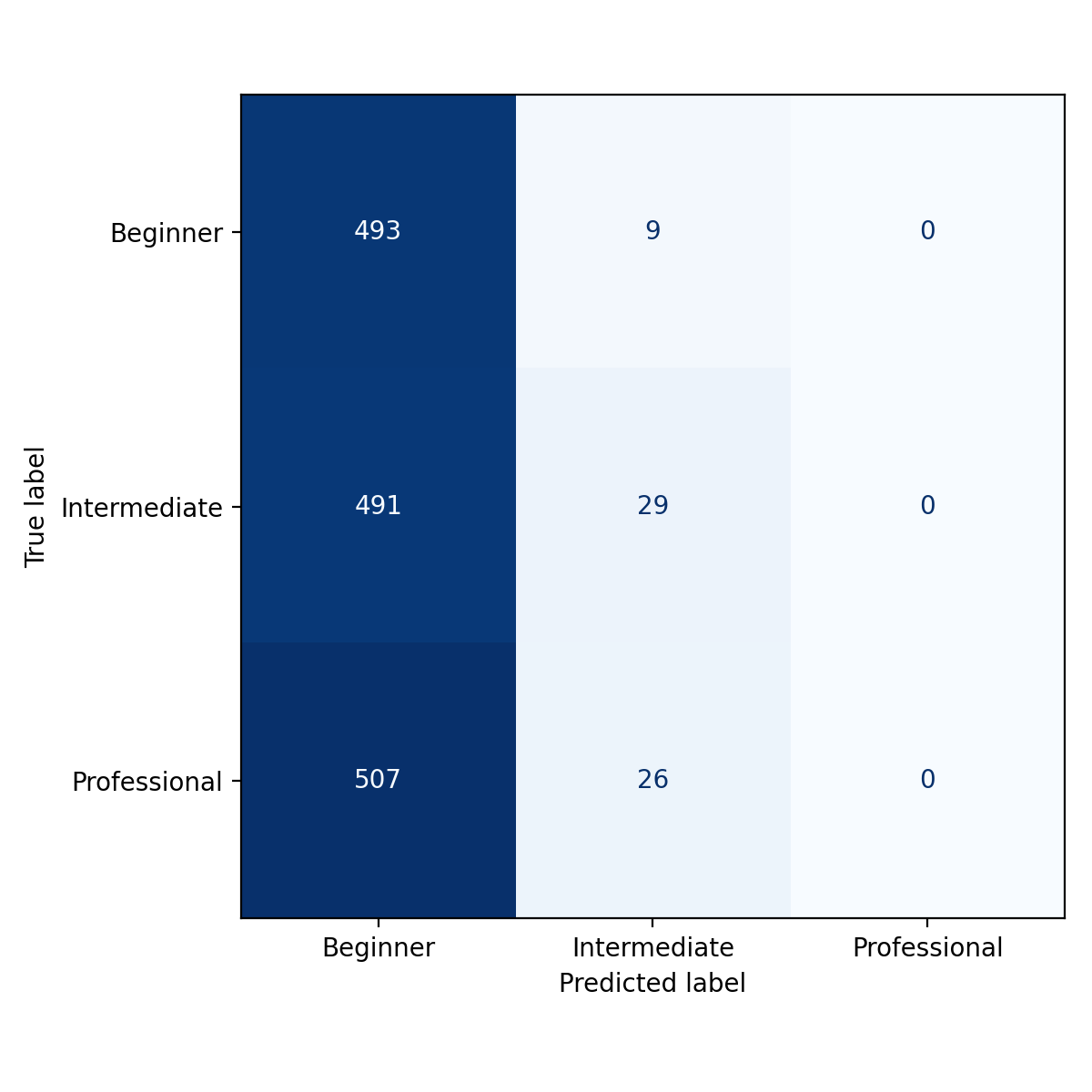}
        \caption{Duolando with zero-shot Qwen2.5-VL.}
    \end{subfigure}
    \vspace{0.5em}

    \caption{Confusion matrices for appropriateness using both zero-shot and fine-tuned Qwen2.5-VL models, reported in Table~\ref{tab:contextual}.}
    \label{fig:cm_appropriateness_level}
\end{figure*}

We evaluate the generated follower motions using our trained classifiers. For legibility, we used the move classifier with the highest F1 score on follower-only videos, which is the fine-tuned InternVL3, and similarly use fine-tuned Qwen2.5-VL for appropriateness (proficiency estimation). To obtain ground truth move labels for the generated clips, we identify the annotated segment with the maximum temporal overlap with the leader segment on which the follower motion was conditioned, and assign its base move label as the ground truth. 

Fig.~\ref{fig:cm_legibility_move} shows the confusion matrices corresponding to the legibility in Table~\ref{tab:contextual}. Notably, the zero-shot model applied to Duolando videos, which achieves the highest accuracy, tends to classify most of the clips as \textit{Basic Step}. Due to class imbalance, this bias inflates accuracy while yielding a low F1 score. In contrast, fine-tuned models predict a more diverse range of labels and achieve consistently higher F1 scores.

For proficiency estimation, we apply the same segmentation procedure described in Section~\ref{appendix-prof-est}. As in all previous experiments, we set \texttt{do\_sample=False} and \texttt{max\_new\_tokens=16} during inference.

Fig.~\ref{fig:cm_appropriateness_level} shows the confusion matrices corresponding to the appropriateness results in Table~\ref{tab:contextual}. As observed in previous settings, zero-shot models tend to over-predict the \textit{Beginner} class, while fine-tuned models produce a more diverse range of predictions, achieving higher accuracy and F1 scores overall.

\clearpage

\raggedbottom

\end{document}